\documentclass{article}

\PassOptionsToPackage{numbers, compress}{natbib}
\usepackage{jmlr2e}

\usepackage[normalem]{ulem}
\usepackage{bbm}
\usepackage[utf8]{inputenc} 
\usepackage[T1]{fontenc}    
\usepackage{hyperref}       
\usepackage{url}            
\usepackage{booktabs}       
\usepackage{amsfonts}       
\usepackage{nicefrac}       
\usepackage{microtype}      
\usepackage{algorithm}
\usepackage{algorithmic}
\usepackage{color}
\usepackage{amsmath}

\usepackage{amsthm}
\usepackage{bbm}
\usepackage[	justification=justified,
   format=plain]{caption}
\newtheorem{asu}{Assumption}

\newtheorem{prop}{Proposition}
\newtheorem{defi}{Definition}
\newtheorem{thm}{Theorem}
\newtheorem{cor}{Corollary}

\newcommand{\ceil}[1]{\lceil #1 \rceil}
\newcommand{\floor}[1]{\lfloor #1 \rfloor}

\begin{document}

\title{An Adaptive Strategy for Active Learning with Smooth Decision Boundary}

\author{\name Andrea Locatelli \email andrea.locatelli@ovgu.de \\
       \addr Department of Mathematics\\
       Otto-von-Güricke Universität Magdeburg\\
       \AND
       \name Alexandra Carpentier \email alexandra.carpentier@ovgu.de\\
       \addr Department of Mathematics\\
       Otto-von-Güricke Universität Magdeburg\\
       \AND
       \name Samory Kpotufe \email samory@princeton.edu \\
       \addr Operations Research and Financial Engineering\\
       Princeton University
       }

\editor{under review for ALT 2018}

\maketitle

\begin{abstract}
We present the first adaptive strategy for active learning in the setting of classification with smooth decision boundary. The problem of adaptivity (to unknown distributional parameters) has remained opened since the seminal work of ~\cite{castro2007minimax}, which first established (active learning) rates for this setting. While some recent advances on this problem establish \emph{adaptive} rates in the case of univariate data, adaptivity in the more practical setting of multivariate data has so far remained elusive. 

Combining insights from various recent works, we show that, for the multivariate case, a careful reduction to univariate-adaptive strategies yield near-optimal rates without prior knowledge of distributional parameters.  


\end{abstract}

\section{Introduction}\label{sec:intro}
In active learning (for classification), the learner can \emph{actively} request $Y$ labels at any point $x$ in the data space to speedup learning: the goal is to return a classifier with low error while requesting as few labels as possible. Previous work (see e.g. \cite{freund1993information,castro2007minimax, Hann2,Kolt,minsker2012a,balcan2009agnostic}) showed that under various distributional settings, active learning offers a significant advantage over passive learning (the usual classification setting with i.i.d. labeled data).

An important such setting is the one studied in the seminal work of~\cite{castro2007minimax}, known as the boundary fragment setting, where the feature space $\mathcal X=[0,1]^d$ is bisected along the $d$-th coordinate by a smooth curve which characterizes the decision boundary $\left\{x: \mathbb{E}[Y|x] = 1/2 \right\}$. The essential error measure in this setting is the distance from the estimated decision boundary to the true decision boundary; such error metric can readily serve to bound the usual $0$-$
1$ classification error under additional distributional assumptions, e.g.,assuming that the marginal $P_X$ is uniform as done in ~\cite{castro2007minimax} (we will relax such assumptions). They show that the minimax optimal rate (in terms of excess $0$-$1$ error over the Bayes classifier) achievable by an active strategy is strictly faster than in the passive setting of ~\cite{tsybakov2004optimal}. While their strategy is minimax optimal, it is unfortunately non-adaptive, i.e., it requires full knowledge of key distributional parameters. Namely, there are two important such parameters: 
$\alpha$, which captures the \textit{smoothness} of the decision boundary, and $\kappa$, which controls the \textit{noise rate}, i.e. \emph{how fast} $\mathbb{E}[Y|x]$ grows away from $1/2$ near the decision boundary. These parameters interpolate between hard and easy problems (rough or smooth decision boundary, high or low noise), and are never known in practice. Therefore, a minimax  adaptive strategy -- i.e., one which attains optimal rates but does not require a priori knowledge of such parameters -- is highly desirable. Such optimal adaptive strategy has unfortunately remained elusive for the general case of data in $\mathbb{R}^d$.

\begin{figure}[t!]
\centering
  \includegraphics[width=0.8\linewidth]{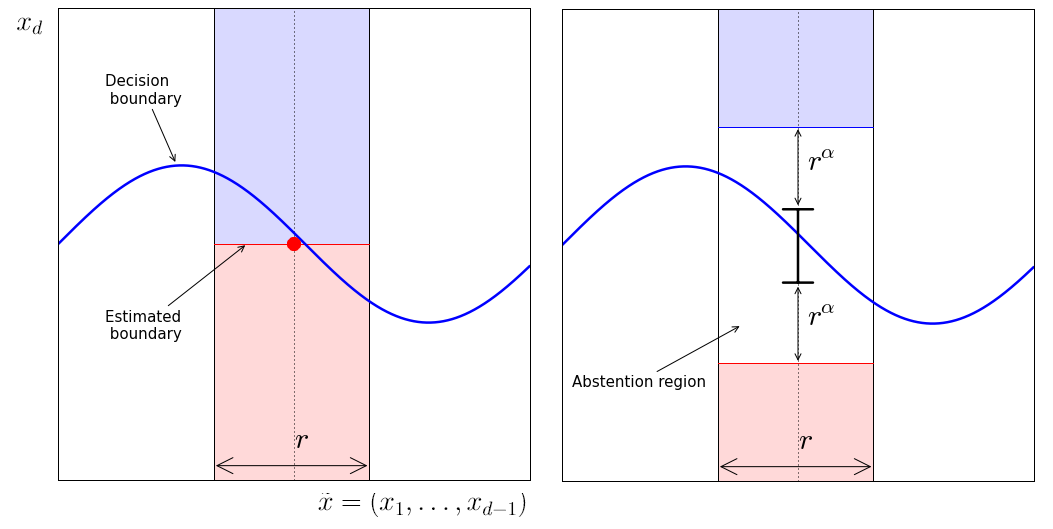}
  \caption{Comparing strategies (for known $\alpha \leq 1$). On the left, we illustrate the strategy in~\cite{castro2007minimax,yan2016active}, and our strategy on the right. Both strategies operate on fixed grids (of $[0, 1]^{d-1}$) with cells of side-length $r$, and perform a line search in each cell (dotted line). On the right, the line search returns a threshold (the red dot) guaranteed to be close to the decision boundary; this threshold is then extrapolated to the entire cell (estimated boundary). The strategy need to operate on an optimal value of $r = r(\alpha, \kappa)$. 
 On the left, the line search returns an interval of size $O(r^\alpha)$, guaranteed to intersect the decision boundary; the interval is then extended by 
 $O(r^\alpha)$ to create an \emph{abstention} region of the right size (in terms of known $\alpha$). To adapt to unknown $\kappa$, the strategy is repeated over dyadic values of $r\to 0$.}
  \label{fig:strat}
\end{figure}

For univariate data ($d = 1$), it is known (\cite{Hann2,ramdas2013algorithmic}) that this limitation can be overcome, and minimax optimal strategies (such as the $A^2$ algorithm in~\cite{balcan2009agnostic}, further studied in ~\cite{hanneke2007bound}) exist, which adapt to unknown noise rate $\kappa$ on the line (there is no notion of smoothness $\alpha$ in the line setting since the boundary is just a threshold). Recently, earlier results of ~\cite{Kolt,hanneke2011rates} -- meant for settings with \textit{bounded disagreement coefficients} -- were extended in ~\cite{wang2011smoothness} to obtain an adaptive procedure for the boundary fragment class of \cite{castro2007minimax}, including the case of data in $\mathbb{R}^d$; unfortunately that strategy yields suboptimal rates for the setting. 

We present the first adaptive and optimal strategy for the setting, by combining insights from various recent work on related problems, and original insights from \cite{castro2007minimax}.

{\bf Combining insights from related work.} The original strategy of \cite{castro2007active} consists of a clever reduction of active learning in $\mathbb{R}^d$ to active learning on $\mathbb{R}$: since the boundary is the curve of function $g:[0, 1]^{d-1} \mapsto [0, 1]$, (a) first partition $[0, 1]^{d-1}$ into a finite number of cells, and do active learning on each cell as follows: (b) pick a line on the cell, and estimate the threshold at which the decision boundary crosses this line; (c) extrapolate the estimated threshold to the whole cell using the fact that the boundary is smooth. Unfortunately step (a) required knowledge of both $\kappa$ and $\alpha$ to pick an optimal cell size, while steps (b) and (c) respectively required knowledge of noise margin $\kappa$ and smoothness $\alpha$. This strategy is illustrated in Figure~\ref{fig:strat} (left box).

A key step in our work, is to temporarily assume knowledge of $\alpha$ and to aim for a procedure that is adaptive to $\kappa$, while following the above strategy of \cite{castro2007minimax}. Clearly, given recent advances on adaptive active learning on $\mathbb{R}$, step (a) above is readily made adaptive to $\kappa$. This is for instance done in the recent work of \cite{yan2016active}, which however leaves open the problem in (a) of choosing a partition of optimal cell-size in terms of unknown $\kappa$ (their work and this issue is discussed in more detail in Section \ref{sec:analysis}). 
We show that we can resolve this issue by proceeding hierarchically over decreasing cell sizes. Furthermore, in order to eventually adapt to unknown $\alpha$, we also require a small but crucial change to the interpolation in step (c) above (the same essential interpolation strategy is used in both \cite{castro2007minimax,yan2016active}. The reason for more careful interpolation is described next. 

In order to adapt to unknown smoothness $\alpha$, we build on recent insights from \cite{locatelli2017adaptivity} which concerns a separate classification setting with smooth regression function $\eta(x) \doteq \mathbb{E} [Y| x]$ rather than smooth decision boundary. Their work presents a generic adaptive strategy that exploits the nested structure of smoothness classes, namely the fact that an $\alpha$-smooth function is also $\alpha'$ -smooth for any $\alpha' < \alpha$. Their strategy consists of aggregating the classification estimates returned by a subroutine taking increasing smoothness values $\alpha'$ as a parameter. The subroutine in our case is that described in the last paragraph -- which takes in the smoothness as a parameter. As it turns out, for the aggregation to work, the subroutine has to be \emph{correct} in a sense that is suitable to our setting, namely, for any $\alpha' < \alpha$, it must only label points that are at an optimal distance away from the decision boundary and \emph{abstain} otherwise (see Figure~\ref{fig:strat}). In other words, the interpolation step (c) discussed above, must produce an \emph{abstention} region of optimal radii in terms of $\kappa$ and $\alpha$. 

Thus, the bulk of our analysis is in constructing a sub-procedure that takes in $\alpha$ as a parameter, is fully adaptive to $\kappa$, and properly abstains in regions of optimal size in terms of $\alpha$ and unknown $\kappa$. Our construction readapts the line-search in \cite{yan2016active} to our particular needs and constraints.

\section{Setting}

In this section, we describe formally the problem of active learning under nonparametric assumptions in the membership query setting.

\subsection{The Active Learning Setting}

{\bf Binary Classification.} We write $\mathbb P_{X, Y}$ for the joint-distribution of feature-label pairs $(X, Y)$. $\mathbb P_X$ denotes the marginal distribution according to variable $X$, supported on $[0,1]^d$. The random variable $Y$ belongs to $\{0, 1\}$ as usual in the binary classification setting. The conditional distribution of $Y$ knowing $X=x$, which we denote $\mathbb{P}_{Y|X=x}$, is characterized by the regression function $$\eta(x) \doteq \mathbb{E}[Y| X = x],~~~\forall x\in [0,1]^d.$$
The Bayes classifier is defined as $f^*(x) = \mathbf 1\{\eta(x) \geq 1/2\}$. It minimizes the $0$-$1$ risk ${R(f) = \mathbb P_{X,Y}(Y \neq f(X))}$ over all possible $f: [0,1]^d \mapsto \{0,1\}$. The aim of the learner is to return a classifier $f$ with small excess error 
\begin{align}
\mathcal{E}(f) \doteq  R(f) - R(f^*) = \int_{x\in [0,1]^d : f(x)\neq f^*(x)} |1 - 2\eta(x)| \text{d} \mathbb P_X(x).
\label{eq:excesserror}
\end{align}

{\bf Active sampling.} At each time $t\leq n$, the active learner can sample a label $Y$ at any $x_t\in \mathbb [0,1]^d$ drawn from the conditional distribution $\mathbb P_{Y | X = x_t}$. 
In total, it can sample at most $n \in \mathbb N^*$ labels - we will refer to $n$ as the \textit{sampling budget} - known to the learner.  At the end of the budget, the active learner returns a classifier $\widehat{f}_n: [0,1]^d \mapsto \{0,1\}$. 

In this work, our goal is to design an adaptive sampling strategy that 
outputs a good estimate of the decision boundary, with high probability over the samples requested and labels revealed, without prior knowledge of distributional parameters, i.e., smoothness and noise margin parameters. This is formalized in Section \ref{sec:dist-params} below.

\subsection{The Nonparametric Setting}
\label{sec:dist-params}
In this section, we expose our assumptions on $\mathbb{P}_{X,Y}$, which are nonparametric in nature, and similar to the setting introduced in~\cite{castro2007minimax}. From now on, we assume that $d \geq 2$.
\begin{defi}[Hölder smoothness] We say that a function $g: [0,1]^{d-1} \mapsto [0,1]$ belongs to the Hölder class $\Sigma(\lambda, \alpha)$ if $g$ is $\floor{\alpha}$\footnote{$\floor{\alpha}$ denotes the largest integer strictly smaller than $\alpha$.} times continuously differentiable and for all $x,y \in [0,1]^{d-1}$, and any $\beta \leq \alpha$ we have:
\begin{equation}
|g(x) - \textsc{TP}_{y,\floor{\beta}}(x)| \leq \lambda ||x-y||_{\infty}^{\beta},
\end{equation}
where $\textsc{TP}_{y,\floor{\beta}}$ is the Taylor polynomial expansion of degree $\floor{\beta}$ of $g$ in $y$ and $||z||_{\infty} \doteq \max_{1 \leq i\leq d} |z_i|$ is the usual infinity norm for $d$ dimensional vectors.

\end{defi}
For any $g \in \Sigma(\lambda, \alpha)$, consider the set $\textrm{epi}(g) \doteq \{ x=(\tilde x, x_d) \in [0,1]^{d-1} \times [0,1]: x_d \geq g(\tilde x) \}$, which is the epigraph of the function $g$. We define the boundary fragment class ${\mathcal G(\lambda, \alpha) \doteq \{\textrm{epi}(g), g \in \Sigma(\lambda, \alpha) \}}$.

\begin{asu}[Smoothness of the boundary]\label{asuH}
There exists constants $\alpha > 0$ and $\lambda \geq 1$ such that $\{x: \eta(x) \geq 1/2 \} \in \mathcal{G}(\lambda, \alpha)$.
\end{asu}
In other words, there exists $g^* \in \Sigma(\lambda, \alpha)$ such that $\{x: \eta(x) \geq 1/2\} = \mathbbm{1}\{\textrm{epi}(g^*)\}$ and the Bayes classifier is equivalent to $\mathbbm{1}\{\textrm{epi}(g^*)\}$. This means that the decision boundary for the classification problem is fully characterized by $g^* \in \Sigma(\lambda, \alpha)$. Importantly, for any $\alpha' \leq \alpha$, we also have $g^* \in \Sigma(\lambda, \alpha')$, as the classes $\Sigma(\lambda, \alpha) \subset \Sigma(\lambda, \alpha')$ are nested for $\lambda$ fixed.\\

We also assume a one-sided noise condition on the behavior of the regression function close to the decision boundary characterized by $g^* \in \Sigma(\lambda, \alpha)$, which can be seen as a geometric variant of the popular Tsybakov noise condition (TNC)(~\cite{tsybakov2004optimal}).

\begin{asu}[Geometric TNC]\label{asuT}
There exists constants $c > 0$ and $\kappa \geq 1$ such that for any ${x = (\tilde x, x_d) \in [0,1]^{d-1} \times [0,1]}$:
\begin{equation*}
|\eta(x) - \frac{1}{2}| \geq c|x_d-g^*(\tilde x)|^{\kappa-1}.
\end{equation*}
\end{asu}

This assumption characterizes how "flat" the regression function $\eta$ is allowed to be in the vicinity of the decision boundary: the larger $\kappa$ the noise parameter, the harder it is to locate the decision boundary precisely. In particular, for $\kappa = 1$, $\eta$ "jumps" at the decision boundary, going from $1/2-c$ to $1/2+c$.\\

In this work, our main objective is to devise an adaptive algorithm that returns an estimate $\hat g$ of the true decision boundary $g^*$, such that $||\hat g - g^*||_\infty$ is small and of optimal size in a minimax sense. Under additional assumptions (which relax original assumptions in~\cite{castro2007active}), we will show that the resulting classifier $x = (\tilde x, x_d) \rightarrow \mathbf 1\{x_d \geq \hat g(\tilde x)\}$ also attains optimal excess risk guarantees.

\begin{defi}
We denote $\mathcal{P}(\alpha, \kappa) \doteq \mathcal{P}(\lambda, \alpha, \kappa, c)$ the set of classification problems $\mathbb P_{X,Y}$ characterized by $(\mathbb P_X, \eta)$ such that Assumption 1 is satisfied for some $g^* \in \Sigma(\lambda, \alpha)$ and Assumption 2 is satisfied with constants $\kappa \geq 1, c> 0$.
\end{defi}

For the rest of the paper we will consider $c>0$ to be fixed, and $\lambda \geq 1$ to be fixed and known to the learner - we discuss the relevance of this assumption in Section~\ref{s:adap}. Now, considering $\kappa$ to be fixed as well as $\lambda$, we remark that the nested structure of the smoothness classes straightforwardly implies the same property for the classes $\mathcal P(\alpha, \kappa)$.

\section{Main Results}

In this section, we show our mains results, assuming we have access to a black-box Subroutine with some correctness property. We first formalize this notion of correctness, and deduct from this a property of the aggregation procedure, which allows us to then state our main adaptive results.

\subsection{Adaptive Algorithm}\label{s:adap}



A first component of our adaptive strategy is a meta-procedure (Algorithm \ref{alg:sa}) that aggregates the classification estimates of a subroutine that takes $\alpha$ as a parameter (but must adapt to unknown noise margin $\kappa$). While much of our analysis concerns this Subroutine, this section introduces the meta-procedure 
whose definition is needed for stating the main result of Theorem \ref{thm_adaptive}. 

The metaprocedure implements original ideas from the recent work of \cite{locatelli2017adaptivity} (which itself adapts ideas in~\cite{lepski1997optimal} to the active setting), which considers a different distributional setting (smoothness of $\eta$ rather than smoothness of the boundary $g^*$) but with a similar nested structure as in this work. The conditions on the Subroutine for the meta-procedure to work in our setting are different, as we will see, and designing a suitable such subroutine constitutes the bulk of our efforts. 



\begin{algorithm}
\caption{Adapting to unknown boundary smoothness $\alpha$}
   \label{alg:sa}
\begin{algorithmic}
   \STATE {\bfseries Input:} $n$, $\delta$, $\lambda$, and a black-box Subroutine
   \STATE {\bfseries Initialization:} $s^{0}_0 =  s^{1}_0 = \emptyset$
   \FOR{$i = 1,..., \floor{\log(n)}^2$}
   \STATE Let $n_0 = \frac{n}{\floor{\log(n)}^2}$, $\delta_0 = \frac{\delta}{\floor{\log(n)}^2}$,  
   and $\alpha_i = \frac{i}{\floor{\log(n)}}$
   \STATE Run Subroutine with parameters $\left(n_0, \delta_0, \alpha_i, \lambda\right)$ and receive $S^{0}_i,S^{1}_i$
   \STATE For $y\in \{0,1\}$, set $s^{y}_i = s^{y}_{i-1} \cup (S^{y}_i \setminus s^{1-y}_{i-1})$
   \ENDFOR
   \STATE {\bfseries Output:} \begin{itemize} 
   
  \item  Confidently labeled sets $S^0 = s^{0}_{\floor{\log(n)}^2}, S^1 = s^{1}_{\floor{\log(n)}^2}$, \\
   \item  Estimated Boundary: $\hat g_n(\tilde x) \doteq \min \{x_d: (\tilde x , x_d) \in S_1 \}$
   \item  Classifier $\hat f_n(x) \doteq \mathbf 1\{x \in S^{1}\} = \mathbf{1}\{x_d \geq \hat g(\tilde x)\}$
   \end{itemize}
\end{algorithmic}
\end{algorithm}

The subroutine is called over increasing guesses $\alpha_i$ of the unknown smoothness parameter $\alpha$ of the boundary, taking advantage of the nested nature of the H\"older classes: if $g^*$ is $\alpha$-H\"older for some unknown $\alpha$, then it is $\alpha_i$-H\"older for $\alpha_i \leq \alpha$. Crucially, the subroutine labels only part of the space, and \emph{abstains} otherwise. Now, suppose that the subroutine, called on $\alpha_i$, guarantees \textit{correctly} labeled sets $S^0_i$, $S^1_i$ whenever $g^*$ is $\alpha_i$-H\"older; then for any $\alpha_i \leq \alpha$ the aggregated labels remain correct. When $\alpha_i > \alpha$, the Subroutine might return incorrect labels. However, this is not a problem since the aggregation procedure never overwrites previously assigned labels, and thus misclassification only occurs in the \emph{abstention} region 
returned by previous calls with $\alpha_i \leq \alpha$. Thus, as long as these abstention regions are of optimal size w.r.t. $\alpha_i \leq \alpha$, the final error of the aggregation procedure will be of optimal order (provided some $\alpha_i \approx \alpha$). 

Following the above intuition, we now formally define \textit{correctness} in a sense suited to our particular setting and implicit goal of estimating the decision boundary. This is different from the notion of \emph{correctness} in \cite{locatelli2017adaptivity} where the goal is to achieve a  correct margin $\Delta$ w.r.t. the regression function $\eta$, i.e. finding $x$ s.t. $|\eta(x) - 1/2| > \Delta$, rather than finding $x$ that are $\Delta$ distant from the boundary $\{x: \eta(x) = 1/2\}$ as in our case. 

\begin{defi}[$(\delta, \Delta,n)$-correct algorithm]\label{def:correct}
Consider a procedure which returns disjoint measurable sets $S^0, S^1 \subset [0,1]^d$. Let $0<\delta<1$, and $\Delta\geq 0$. 
We call such a procedure {\bf weakly} $(\delta, \Delta,n)$-{\bf correct} for a classification problem $\mathbb P_{X,Y} \in \mathcal P(\alpha, \kappa)$ if, with probability larger than $1-2\delta$ using at most $n$ label requests: 
\begin{align*} 
&\left \{x=(\tilde x, x_d) \in [0,1]^d : x_d - g^*(\tilde x) > \Delta \right \} \subset S^1\\
&\left\{x=(\tilde x, x_d) \in [0,1]^d : g^*(\tilde x) - x_d > \Delta\right\} \subset S^0.
\end{align*}
If in addition, under the same probability event over at most $n$ label requests, we have  
\begin{align*} 
S^1 \subset&\left \{x=(\tilde x, x_d) \in [0,1]^d : x_d > g^*(\tilde x) \right \}\\
S^0\subset &\left \{x=(\tilde x, x_d) \in [0,1]^d : x_d < g^*(\tilde x) \right \}
\end{align*}
then such a procedure is simply called $(\delta, \Delta,n)$-{\bf correct} for $\mathbb P_{X,Y}$.
\end{defi}

In the boundary fragment setting, correctness is defined in terms of distance to the decision boundary, which is a major difference with respect to the smooth regression function (see~\cite{locatelli2017adaptivity} and the different notion of correctness therein). Importantly, a correct procedure returns labeled sets with the following key properties (with high probability): first, points are always labeled in agreement with their true class (and thus, bring no excess risk). Second, it abstains in a region of width at most $\Delta$ around the true decision boundary.

\subsection{Main Results}
In this Section we present our main result,  Theorem~\ref{thm:adap_rate}, which bounds the distance from our estimated boundary to the true boundary. As a corollary, the excess $0$-$1$ risk of the estimated classifier can be bounded under additional distributional assumptions that relax the original setting of \cite{castro2007active}. 

We start with the following simple proposition, stating (as in the intuition detailed above) that Algorithm~\ref{alg:sa} correctly aggregates estimates whenever the subroutine calls return correct estimates.\\

\begin{prop}[Correctness of aggregation]\label{thm_adaptive}
Let $n \in \mathbb N^*$ and $1>\delta>0$.  Let $\delta_0 = \delta/(\floor{\log(n)}^2)$ and $n_0 = n/(\floor{\log(n)}^2)$ as in Algorithm~\ref{alg:sa}. Fix $\kappa\geq 1$. Suppose that, for any $\alpha >0$,   
the Subroutine in Algorithm~\ref{alg:sa} is $(\delta_0, \Delta_{\alpha},n_0)$-correct for any $\mathbb P_{X,Y} \in \mathcal P(\alpha, \kappa)$, where $\Delta_{\alpha}>0$ depends on $n,\delta$ and the class $\mathcal P(\alpha, \kappa)$.

Fix $\alpha \in [\floor{\log(n)}^{-1}, \floor{\log(n)}]$, and let $\alpha_i = i/\floor{\log(n)}$ for $i \in \{1, \ldots, \floor{\log(n)}^2\}$. Then Algorithm~\ref{alg:sa} is {\bf weakly} $(\delta_0, \Delta_{\alpha_i}, n_0)$-correct for any $\mathbb P_{X,Y} \in \mathcal P(\alpha, \kappa)$  for the largest $i$ such that $\alpha_i \leq \alpha$. 
\end{prop}
The proof of this proposition follows can be found in Section~\ref{proof:adap_rate} of the Appendix, and follows from arguments in~\cite{locatelli2017adaptivity}. The main difference in the interpretation of this result with respect to the result in~\cite{locatelli2017adaptivity}, in which correctness is defined in terms of distance between $\eta$ and $1/2$, is that we are interested here in locating the decision boundary $g^*$. This makes Proposition~\ref{thm_adaptive} very simple to visualize in our setting. For any run with $\alpha_i \leq \alpha$, the decision boundary is estimated within a margin $\Delta_{\alpha_i}$ such that no regions are misclassified. As $\alpha_i \leq \alpha$ grows, this margin decreases, until it reaches the largest $i^*$ such that $\alpha_{i^*} \leq \alpha$. For any $i > i^*$, we cannot characterize the behavior of the non-adaptive Subroutine; fortunately, the misclassified regions are confined to the set $\left \{x=(\tilde x, x_d) \in [0,1]^d : |x_d - g^*(x)| < \Delta_{\alpha_{i^*}} \right \}$.\\

We now state our main adaptive result (Theorem \ref{thm:adap_rate}). Following Proposition~\ref{thm_adaptive}, the main work in obtaining Theorem \ref{thm:adap_rate} consist of producing a Subroutine that is \emph{correct} in the sense of Definition~\ref{def:correct}. This is done in Theorem~\ref{thm:kappa} of Section \ref{sec:analysis}. 

\begin{thm}\label{thm:adap_rate}
Let $n \in \mathbb N^*$ and $\delta > 0$. Assume that $\mathbb P_{X,Y} \in \mathcal P(\alpha, \kappa)$ with $\alpha \in [\floor{\log(n)}^{-1}, \floor{\log(n)}]$. Algorithm~\ref{alg:sa} run with parameters $(n, \delta, \lambda)$ and using Algorithm~\ref{alg:Subroutine2} as the black-box Subroutine outputs an approximation of the decision boundary $\hat g_n$ such that with probability at least $1-2\delta$:
\begin{equation*}
\mathcal ||\hat{g}_n-g^*||_\infty \leq C \lambda^{\frac{(d-1)}{2\alpha(\kappa-1)+d-1}}\left( \frac{\log^3(n/\delta)}{n} \right)^{\alpha/(2\alpha(\kappa-1)+d-1)},
\end{equation*}
where $C>0$ is a constant that does not depend on $\lambda, n, \delta$.
\end{thm}

The proof of this Theorem can be found in Section~\ref{proof:adap_rate}.

By setting $\delta = n^{-\log(n)/(d-1)}$ in Theorem~\ref{thm:adap_rate}, we also get a rate in expectation of order $\tilde O\left(n^{-\alpha/(2\alpha(\kappa-1)+d-1)}\right)$, matching (up to logarithmic factors) the minimax lower bound derived in~\cite{castro2007minimax}, despite not having access to $\alpha$ nor $\kappa$.\\

So far, we have made no assumption on $\mathbb P_X$. In order to relate this bound on the distance between $\hat g_n$ and $g^*$ to a guarantee on the risk of the classifier $\hat f_n$, we now state a third assumption, which bounds the risk incurred by regions that are close to $g^*$.

\begin{asu}\label{asuM}
There exists $C > 0$, $\Delta_0 > 0$ and $\kappa' > 0$ such that $\forall \Delta \in [0, \Delta_0]$:
\begin{align*}
\int_{x\in [0,1]^d : |x_d - g^*(\tilde x)| \leq \Delta} |1 - 2\eta(x)| \text{\emph{d}}\mathbb P_X(x) \leq C \Delta^{\kappa'}
\end{align*}
\end{asu}

This assumption relaxes the setting introduced in~\cite{castro2007minimax}, as we will see in Example 1 (in particular there, $\kappa' = \kappa$ which can be strong). Assumption~\ref{asuM} and Theorem~\ref{thm:adap_rate} directly lead to the following corollary, which bounds the excess risk of the classifier with high probability.

\begin{cor}\label{cor:kappa_p}
Under the assumptions of Theorem~\ref{thm:adap_rate} and Assumption~\ref{asuM}, for $n\geq N = N(\alpha, \lambda, \kappa, \delta)$, Algorithm~\ref{alg:sa} run with $(n, \delta, \lambda)$ outputs a classifier $\hat f_n$ such that with probability at least $1-2\delta$ its excess risk is bounded as:
\begin{equation*}
\mathcal E(\hat{f}_n)\leq C \lambda^{\frac{\kappa'(d-1)}{2\alpha(\kappa-1)+d-1}}\left( \frac{\log^3(n/\delta)}{n} \right)^{\alpha \kappa'/(2\alpha(\kappa-1)+d-1)},
\end{equation*}
where $C>0$ is a constant that does not depend on $\lambda, n, \delta$.
\end{cor}

From the corollary we see that larger values of $\kappa'$ and lower values for $\kappa$ improve the rate; this can be a source of tension under the restriction that $\kappa' = \kappa$ as in the first example below. The first example below is the exact setting of \cite{castro2007minimax}.  

\textbf{Example 1} (\cite{castro2007minimax}).
Consider $\mathbb P_X$ uniform over $[0,1]^d$ and $\eta$ such that: 
$$c|x_d - g^*(\tilde x)|^{\kappa-1} \leq \left |\eta(x)-\frac{1}{2}\right| \leq C|x_d - g^*(\tilde x)|^{\kappa-1}.$$ 
It is clear that Assumption~\ref{asuM} is satisfied with $\kappa' = \kappa$. Under these assumptions, the minimax rate in expectation for the excess risk is of order $\Omega(n^{-\alpha\kappa/(2\alpha(\kappa-1)+d-1)})$ as shown by~\cite{castro2007minimax}. Our procedure is the first adaptive and optimal (up to logarithmic factors) strategy in this setting. Notice that in this case, both low and large values of $\kappa$ seem to improve the rate. 
In fact, for $\alpha> (d-1)/2$ we get \emph{fast rates} (below $n^{-1/2}$) and lower values of $\kappa$ improve the rate. On other hand, when $\alpha \leq (d-1)/2$, greater values of $\kappa$ improve the rate. This tension comes from the fact that lower values of $\kappa$ on the one hand make it easier to locate the decision boundary as there is a sharper jump close to $g^*$; yet for large values of $\kappa = \kappa'$, misclassifying a large region close to the boundary bears less risk. Assumption \ref{asuM} decouples the effect of $\kappa$ and $\kappa'$, which is evident in the following example.\\

\textbf{Example 2} (Hard and soft margin in $\mathbb P_X$). Consider situations where $\mathbb P_X$ has little or no mass near the decision boundary. First consider the extreme of no mass near the boundary (hards margin), i.e. there exists $\Delta_0$ such that 
$$ \mathbb P_X(x : |x_d - g^*(\tilde x)| \leq \Delta_0) = 0 .$$
In this case $\kappa' = \infty$ in Assumption~\ref{asuM}, and the classifier attains $0$ error with high probability (equivalently, exponentially small error in expectation). More generally (soft-margin) Assumption \ref{asuM} holds if $\mathbb P_X$ decreases sufficiently fast near the boundary: for instance, suppose we have $\forall\,  0< \Delta \leq \Delta_0$, 
$$ \mathbb P_X(x : |x_d - g^*(\tilde x)| \leq \Delta) \leq \Delta^{\kappa' - \kappa_0 +1},$$
where $\kappa_0 \leq \kappa\wedge (\kappa'+1)$ satisfies the upper-bound $\left |\eta(x)-\frac{1}{2}\right| \leq c|x_d - g^*(\tilde x)|^{\kappa_0-1}$.


We complete this result with the following lower bound, which shows that the rate in Corollary~\ref{cor:kappa_p} is tight up to logarithmic factors, at least for $\kappa' > \kappa-1$, and strictly faster than the passive rate under the same assumptions.
\begin{thm}[Active Lower Bound]\label{thm:lb}
Let $\alpha > 0, \kappa > 1$ and $\kappa' > \kappa-1$. Consider $\mathcal P(\alpha, \kappa, \kappa')$ the subset of $\mathcal P(\alpha, \kappa)$ such that Assumption~\ref{asuM} is satisfied with $\kappa'$. For $n$ large enough, any (possibly active) strategy $\mathcal A_n$ that collects at most $n$ samples before returning a classifier $\widehat{f}_n$ satisfies:
$$
\inf_{\mathcal A_n} \sup_{\mathbb P_{X,Y} \in \mathcal{P}(\alpha, \kappa, \kappa')} \mathbb{E}[\mathcal{E}(\widehat{f}_n)] \geq C n^{-\alpha\kappa'/(2\alpha(\kappa-1)+d-1)},
$$
where $C>0$ does not depend on $n$ and the expectation is taken with respect to both the samples collected by the strategy $\mathcal A_n$ and $\mathbb P_{X,Y}$.
\end{thm}

Finally, we derive a lower bound in the passive setting, 
in terms of $\kappa'$ (previous lower-bounds for related settings do not consider $\kappa'$, see for example \cite{tsybakov2004optimal}). The lower-bound below highlights the gains in active learning, as the rate of Corollary \ref{cor:kappa_p} and Theorem \ref{thm:lb} is strictly faster than the passive-learning lower-bound obtained below.

\begin{thm}[Passive Lower Bound]\label{thm:lb_passive}
Under the Assumption of Theorem~\ref{thm:lb}, for $n$ large enough, any classifier $\widehat{f}_n$ trained on at most $n$ i.i.d. samples satisfies:
$$
\inf_{ \widehat{f}_n} \sup_{\mathbb P_{X,Y} \in \mathcal{P}(\alpha, \kappa, \kappa')} \mathbb{E}[\mathcal{E}(\widehat{f}_n)] \geq C n^{-\alpha\kappa'/(\alpha(\kappa+\kappa'-1)+d-1)},
$$
where $C>0$ does not depend on $n$.
\end{thm}

For $\kappa = \kappa'$, the lower-bound recovers known rates for passive learning see e.g. \cite{tsybakov2004optimal, castro2007active}. 

The proofs of these Theorems can be found in the Section~\ref{proof:lb} and~\ref{proof:lb_passive} of the Appendix. It is based on  general information theoretic arguments (Fano's method) as exposed in a suitable form by~\cite{tsybakovintroduction} and adapted to active learning by~\cite{castro2007minimax}. The geometric construction builds on lower-bound constructions in ~\cite{locatelli2017adaptivity} for the separate setting of smooth regression functions.

\section{Analysis}
\label{sec:analysis}
\subsection{A $\kappa$-Adaptive Procedure for the Boundary Fragment Class} 
We now introduce an algorithm that is fully adaptive with respect to $\kappa$ the noise parameter, and takes as input $\alpha$, $\lambda$ the smoothness parameters of the decision boundary such that $g^* \in \Sigma(\lambda, \alpha)$. The strategy uses as a subroutine another adaptive procedure that solves the unidimensional problem of finding a threshold $x_d^*$ such that for $\tilde x \in [0,1]^{d-1}$ fixed $g^*(\tilde x) = x_d^*$, we will refer to this univariate problem as the \textit{line-search} problem in our context. In this section, we assume that we have access to a line-search procedure such that when it is called with a certain confidence $\delta$ and precision $\epsilon$, it returns a threshold estimate $T$ such that $|T-x_d^*| \leq \epsilon$ with probability at least $1-\delta$ using at most $\tilde{\mathcal O}(\epsilon^{-2(\kappa-1)})$ samples\footnote{we use $\tilde{\mathcal O}$ to hide logarithmic factors in $\frac{1}{\delta}$ and $\frac{1}{\epsilon}$}. Such a procedure was proposed in the recent work of~\cite{yan2016active}. In their work, they use this procedure as a subroutine in the setting where one wants to estimate the boundary with a $\hat g$ such that $||\hat g - g^*|| \leq \epsilon$ with high probability. Assuming knowledge of the smoothness $\alpha$ and given a target error $\epsilon$, they can guarantee a number $n = n(\epsilon)$ of label requests optimal and adaptive in terms of unknown $\kappa$. Interestingly, given the goal of fixed target error $\epsilon$, the problem of adaptive cell size as exposed in Section~\ref{sec:intro} seems to disappear: it's sufficient to partition $[0, 1]^{d-1}$ into cells of size $\epsilon^{1/\alpha}$. The procedure they use is the same as the one exposed in~\cite{castro2007minimax}, as both strategies rely on a discretization of $[0,1]^{d-1}$, launch a number of line-searches on a grid that covers the feature space, and then use the threshold estimates on this grid to construct a smooth approximation of the boundary such that $||g^*-g^*||\leq \epsilon$. However, in our setting (and that of \cite{castro2007active}) we instead fix a labeling budget $n$ and aim to achieve an error $\epsilon = \epsilon(n)$ adaptive to unknown $\kappa$; in other words, to use the algorithmic strategy of \cite{yan2016active} we need knowledge of the optimal $\epsilon$ (which depend on unknown $\kappa$) in order to define an optimal partition cell size. Indeed, in this fixed budget setting, the strategy in~\cite{castro2007minimax} uses both $\kappa$ and $\alpha$ to find the right step-size for the discretization which is of order $\floor{n^{-\alpha/(2\alpha(\kappa-1)+d-1)}}$. Our strategy bypasses this issue by proceeding hierarchically over a dyadic partition of 
$[0,1]^{d-1}$. Our stopping criterion for the line-search procedure only depends on $\alpha, \lambda$ and the cell size, and allows our procedure to fully adapt to $\kappa$. As a last step, we carefully select the regions to label -- and hence the abstention region -- so as to make the procedure \emph{correct} in the sense of Definition~\ref{def:correct}.


\begin{algorithm}
\caption{$\kappa$-adaptive procedure in $d$-dimension}
   \label{alg:Subroutine2}
\begin{algorithmic}
	\STATE \textbf{Input:} $n$, $\delta$, $\lambda$, $\alpha$
    \STATE \textbf{Initialisation:} $l=1$, $t=0$
      \WHILE{$\{t < n\}$}
    \STATE $M_l = \max(1, \floor{\alpha})2^{l}$
    \STATE $\epsilon_l = \lambda 2^{-l\alpha}$
    \STATE $\delta_l = \delta  (\max(1,\floor{\alpha})2^{l(d+1)})^{-1}$
    \FOR{each $\tilde a$ in $\{0,...,M_l\}^{d-1}$}
    \STATE Run Subroutine~\ref{alg:Subroutine} on the line $\mathcal L_{\tilde a}$ with parameters $\epsilon_l$, $\delta_l$
    \STATE Receive threshold estimate $T_{l, \tilde a}$ and budget used $N_{l,\tilde a}$
   \ENDFOR
   \STATE Compute total budget used at depth $l$: $N_l = \sum_{\tilde a} N_{l,\tilde a}$
   \STATE $t = t + N_l$
   \STATE $l = l+1$
   \ENDWHILE
   \STATE $l^* \doteq l-1$ (final completed depth)
   \STATE Fit $\lfloor \alpha \rfloor$-degree tensor-product Lagrange polynomial approximation of boundary using $(T_{l^*,\tilde a})_{\tilde a}$
   \STATE $b_{l^*} \doteq \lambda \ceil{\alpha}^{d\ceil{\alpha}} M_{l^*}^{-\alpha}$ (bias term)
   \STATE $S^0 \doteq \{x: x_d \leq \widehat{P}(\tilde x) - 4 b_{l^*}\}$
   \STATE $S^1 \doteq \{x: x_d \geq \widehat{P}(\tilde x) + 4 b_{l^*}\}$
   \STATE \textbf{Output:} $S^y$ for $y \in \{0,1 \}$
\end{algorithmic}
\end{algorithm}

Our procedure, Algorithm~\ref{alg:Subroutine2}, takes as input $n$ the maximum sampling budget, $\delta$ a confidence parameter, as well as $\lambda$ and $\alpha$ the smoothness parameters such that $g^* \in \Sigma(\lambda,\alpha)$. At each depth $l$, the algorithm launches $(M_l+1)^{d-1}$ line-searches with $M_l = \max(1, \floor{\alpha})2^l$, on a grid of step $M_l^{-1}$. Precisely, for each $\tilde a \in \{0, ..., M_l\}^{d-1}$ it launches a line-search instance using Algorithm~\ref{alg:Subroutine} on the line segment ${\mathcal L_{\tilde a} \doteq \{(M_l^{-1}\tilde a, x_d), x_d \in [0,1] \}}$ with confidence parameter $\delta_l = \delta  (\max(1,\floor{\alpha})2^{l(d+1)})^{-1}$ and precision  $\epsilon_l = \lambda 2^{-l\alpha}$. Importantly, the precision with which the line-search procedure is called depends only on the step-size of the grid and the smoothness parameters $\lambda$ and $\alpha$, and not on $\kappa$. Heuristically, the precision of the line-search need not be greater than the precision of the nonparametric approximation of degree $\floor{\alpha}$ of the boundary fit with the estimated thresholds on the grid of step size $M_l^{-1}$, which motivates our choice for $\epsilon_l$. After each run indexed by $\tilde a$, it receives the estimated threshold $T_{l, \tilde a}$ and the budget used $N_{l, \tilde a}$. While the total budget used is less than the maximum allowed budget $n$, the discretization is refined and line-searches are initialized with a higher precision parameter. Once the budget has run out, we use the estimated thresholds $(T_{l^*,\tilde a})_{\tilde a}$ at the last depth $l^*$ such that all the line-searches have terminated to construct a polynomial interpolation of degree $\floor{\alpha}$ of the boundary, as in the original strategy of~\cite{castro2007minimax}. In the case of $\alpha \leq 1$, we simply use in each cell a constant approximation that takes the value of the estimates $(T_{l^*,\tilde a})_{\tilde a}$, the details of which can be found in the proof of Theorem~\ref{thm:kappa}. In what follows, we assume $\alpha > 1$ and describe the approximation method for higher order smoothness.

To that effect, we will use the tensor-product Lagrange polynomials as in~\cite{castro2007minimax} on slightly larger cells, to ensure that the number of estimated thresholds (coming form the line-searches) in those cells is enough to fit a $\floor{\alpha}$-degree polynomial approximation. Let $\tilde q \in \{0, ..., \frac{M_{l^*}}{\floor{\alpha}}-1 \}^{d-1}$ index the cells:
\begin{equation*}
I_{\tilde q} \doteq \Big[\tilde q_1 \floor{\alpha}M_{l^*}^{-1}, (\tilde q_1 +1)\floor{\alpha}M_{l^*}^{-1}\Big] \times ... \times \Big[\tilde q_{d-1} \floor{\alpha}M_{l^*}^{-1}, (\tilde q_{d-1} +1)\floor{\alpha}M_{l^*}^{-1}\Big].
\end{equation*}
These cells partition $[0,1]^{d-1}$ entirely, as we have $M_{l^*}=\floor{\alpha}2^{l^*}$. We use the tensor-product Lagrange polynomial basis as in~\cite{castro2007minimax}, defined as follows:
\begin{equation*}
Q_{\tilde q, \tilde a}(\tilde x) \doteq \prod_{i=1}^{d-1} \ \prod_{\substack{ 0 \leq j \leq \floor{\alpha} \\ j \neq \tilde a_i - \floor{\alpha} \tilde q_i}} \frac{\tilde x_i-M_{l^*}^{-1}(\floor{\alpha}\tilde q_i+j)}{M_{l^*}^{-1}\tilde a_i - M_{l^*}^{-1}(\floor{\alpha}\tilde q_i+j)}.
\end{equation*}
Importantly, this polynomial basis has the following property $\max_{\tilde x \in I_{\tilde q}} |Q_{\tilde a, \tilde q}(\tilde x)| \leq \floor{\alpha}^{(d-1)\floor{\alpha}}$. We define the estimated polynomial interpolation of $g^*$ for $\tilde x \in I_{\tilde q}$:
\begin{equation*}
\widehat{P}_{\tilde q}(\tilde x) \doteq \sum_{\substack{\tilde a \in \{0,..,M_{l^*}\}^{d-1} \\ \tilde a: M_{l^*}^{-1}\tilde a \in I_{\tilde q}}} T_{l^*, \tilde a}Q_{\tilde q, \tilde a}(\tilde x).
\end{equation*}
This polynomial interpolation scheme is such that for any $\tilde a \in \{0,..,M_{l^*}\}^{d-1}$ with $M_{l^*}^{-1}\tilde a \in I_{\tilde q}$, we have $\widehat{P}_{\tilde q}(M_{l^*}^{-1} \tilde a) = T_{l^*,\tilde a}$ i.e. we can control exactly the value of the interpolation on the grid. We also define for the entire feature space: $\widehat{P}(\tilde x) \doteq \sum_{\tilde q} \widehat{P}_{\tilde q}(\tilde x)\mathbf{1}\{\tilde x\in I_{\tilde q}\}$.

Finally, we define ${b_{l^*} = \lambda \ceil{\alpha}^{d\ceil{\alpha}} M_{l^*}^{-\alpha}}$ which is a \textit{bias term} related to the interpolation method we use. Points that are far away enough from the estimate $\widehat{P}$ of the boundary with respect to this bias term are then labeled by the algorithm, as we assign $S^0 \doteq \{x: x_d \leq \widehat{P}(\tilde x) - 4 b_{l^*}\}$ and $S^1 \doteq \{x: x_d \geq \widehat{P}(\tilde x) + 4 b_{l^*}\}$ to the labels $0$ and $1$ respectively. This careful labeling is crucial for the Subroutine to have the desired properties to be used in the aggregation procedure.

The following theorem shows that Algorithm~\ref{alg:Subroutine2} is an acceptable subroutine for the adaptive procedure, as it is \textit{correct} in the sense of Definition~\ref{def:correct}.
\begin{thm}\label{thm:kappa}
Algorithm~\ref{alg:Subroutine2} run on a problem in $\mathcal P(\alpha, \kappa)$ with parameters $n, \delta, \lambda, \alpha$ is $(\delta, \Delta_n, n)$-correct with $\Delta_n$ such that:
\begin{equation*}
\Delta_n \leq 7\ceil{\alpha}^{d\ceil{\alpha}} 2^\alpha \lambda^{\frac{d-1}{2\alpha(\kappa-1)+d-1}} \Big(\frac{\log(n/\delta)}{c_1n} \Big)^{\frac{\alpha}{2\alpha(\kappa-1)+d-1}},
\end{equation*}
where $c_1 = \frac{(\kappa-1)c^2}{400\left(2\ceil{\alpha}\right)^{d-1} \alpha \log(1/c) \kappa 8^{2(\kappa-1)}}$, and where $c$ is the constant involved in Assumption~\ref{asuT}.
\end{thm}
The proof of this result can be found in the Appendix in Section~\ref{thm:kappa}.

\subsection{Learning One-Dimensional Thresholds}\label{adap:1D}
In this section, we briefly describe the procedure (derived from recent advances in~\cite{yan2016active}) whose objective is to actively find a threshold in the one dimensional problem (see~\cite{castro2006upper,Hann2,ramdas2013algorithmic}). This procedure, Algorithm~\ref{alg:Subroutine} is adaptive with respect to $\kappa$, and is used as a Subroutine for the more involved $d$ dimensional procedure. Fix $\tilde x \in [0,1]^{d-1}$, and assume that there exists $g^*$ such that $\eta$ satisfies Assumptions~\ref{asuH} and~\ref{asuT}. In the line-search problem, the goal is to find $x^*= (\tilde x, x_d^*)$ such that $g^*(\tilde x) = x_d^*$, which is equivalent to finding $x_d^*$ such that $\eta(x^*) \geq 1/2$ and for any $x_d < x_d^*$, $\eta((\tilde x, x_d)) < 1/2$. The objective of the Subroutine is to return an interval of length at most $\epsilon$ such that the threshold $x_d^*$ is contained in this interval with probability at least $1-\delta$ and with optimal sample complexity $N$.\\

\begin{algorithm}
\caption{Univariate $\kappa$-adaptive procedure (line-search) - \cite{yan2016active}}
   \label{alg:Subroutine}
\begin{algorithmic}
	\STATE \textbf{Input:} $\epsilon$, $\delta$
    \STATE \textbf{Initialisation:} $[L_1, R_1] \gets [0,1]$, $k=0$, $N=0$, $K = \lceil \log_2(\frac{1}{2\epsilon}) \rceil$
      \WHILE{$\{k< K-1\}$}
    \STATE $k \gets k+1$, $t_k = 0$, $\delta_k = \frac{\delta}{K2^k}$
   	\STATE $M_k \gets \frac{L_k + R_k}{2}$, $U_k \gets \frac{R_k - L_k}{4}+L_k$, $V_k \gets \frac{R_k - L_k}{4}+M_k$
    \WHILE{\TRUE}
    \STATE $t_k = t_k+1$; $N = N+3$
    \STATE Request labels in $M_k, U_k, V_k$, receive $Y_{t_k}(M_k), Y_{t_k}(U_k), Y_{t_k}(V_k)$
    \STATE Estimate $\eta$ for $Z \in \{U_k, M_k, V_k\}$: $\widehat \eta_{t_k}(Z) = t_k^{-1} \sum_{i=1}^{t_k} Y_i(Z)$
    \IF{$|\widehat{\eta}_{t_k}(M_k) - 1/2| \geq 2\sqrt{\frac{\log(t_k/\delta_k)}{2t_k}}$}
    	\IF{$\widehat{\eta}_{t_k}(M_k)>1/2$}
        	\STATE $[L_{k+1},R_{k+1}] \gets [L_k, M_k]$; \textbf{break}
         \ELSE
         	\STATE $[L_{k+1}, R_{k+1}] \gets [M_k, R_k]$; \textbf{break}
         \ENDIF
    \ENDIF
    \IF{$\widehat{\eta}_{t_k}(V_k) - 1/2 \geq 2\sqrt{\frac{\log(t_k/\delta_k)}{2t_k}}$ \textbf{and} $1/2 - \widehat{\eta}_{t_k}(U_k) \leq 2\sqrt{\frac{\log(t_k/\delta_k)}{2t_k}}$}
    \STATE $[L_{k+1}, R_{k+1}] \gets [U_k,V_k]$; \textbf{break}
    \ENDIF
    \ENDWHILE
   \ENDWHILE
   \STATE \textbf{Output:} $L_K, R_K, T_K = \frac{R_K - L_K}{2}$ (threshold estimate), $N \leq n$ (budget used)
\end{algorithmic}
\end{algorithm}

The procedure we use is a natural adaptation of the famous bisection method for root-finding of deterministic monotone functions in one-dimension, and is a simplification of the strategy in~\cite{yan2016active}, as we do not allow abstention for the labeling oracle. In the deterministic setting, a simple strategy is to query the middle point of the active segment, and depending on the label returned by the query, continue the procedure with one of the two subintervals - effectively dividing by two the length of the active region with each epoch. In the stochastic setting, the intuition is similar, however, at epoch $k$, we query successively three active points - the three quartiles of the active segment $[L_k, R_k]$, until we know with a certain confidence $\delta_k$ the label of some of these active points. This is done by comparing the empirical mean of the labels observed in each point, with the threshold $1/2$ and a confidence term that depends on the number of times we have queried the active points. If we know with a certain confidence the label of $M_k$, the median of $[L_k, R_k]$, we simply start the next epoch following the strategy in the deterministic setting, and the active segment is divided by a factor of $2$. However, $M_k$ can be arbitrarily close to the threshold $x^*$, and $\eta(M_k)$ arbitrarily close to $1/2$ (in the case $\kappa > 1$), making it possibly very difficult to discover its true label - which is why we also query two other points $U_k$ and $V_k$. Similarly, if we know with high probability the labels of $U_k$ and $V_k$, the next epoch $k+1$ is started on the segment $[U_k, V_k]$, which also divides the length of the active segment by $2$. The algorithm terminates when it reaches the depth $K = \ceil{\log_2(\frac{1}{2\epsilon})}$ and outputs a final threshold estimate $T_K$ and $N$ the total labeling budget used. The following theorem gives a bound on the number of samples required to return an interval of length at most $\epsilon$ such that with high probability the true threshold $x^*_d$ is in this interval.

\begin{thm}\label{thm:ls_bound}
Fix $\tilde x \in [0,1]^{d-1}$ and let $x_d^* = g^*(\tilde x)$ and assume that $g^*$ and $\eta$ satisfy Assumption~\ref{asuT}. Algorithm~\ref{alg:Subroutine} run with precision $\epsilon$ and confidence $\delta$ terminates with probability at least $1-\delta$ and returns a threshold estimate $T_K$  such that $|T_K - x_d^*| \leq \epsilon$ using at most $N$ samples with
\begin{equation*}
N \leq \begin{cases}
64 \big(\log(\frac{1}{\delta})+\log(\frac{1}{\epsilon})\big)\frac{\log(1/c)}{c^{2}}\log(\frac{1}{\epsilon}), \ \text{if} \ \ \kappa = 1 \\
\quad \\
200\Big(\log(\frac{1}{\delta})+\log(\frac{1}{\epsilon})\Big) \frac{\kappa \log(c^{-1}) 8^{2(\kappa-1)}}{(\kappa-1)c^2}\Big(\big(\frac{1}{\epsilon}\big)^{2(\kappa-1)}-1\Big), \ \text{if} \ \ \kappa > 1,
\end{cases}
\end{equation*}
where $c$ is the constant involved in Assumption~\ref{asuT}.
\end{thm}

\subsection{Remarks on the Subroutines}
Both Subroutines make optimistic guesses on the labels of the queried points, inspired from techniques in the bandit literature (in particular UCB-based strategies \cite{auer2002finite} - see~\cite{bubeck2012regret} for a survey). In the classification setting, the quantity of interest for a point $x$ is how far this point is from the decision boundary $g^*(x)$, or how far $\eta(x)$ is from $1/2$. By using a confidence term, it is possible to determine with a certain confidence the label of $x$, or avoid making a potentially wrong guess. In our setting, this observation naturally leads to efficient algorithms that are able to find the decision boundary (up to a certain precision). These optimistic guesses are crucial to show the correctness property required by the aggregation strategy adapted from~\cite{locatelli2017adaptivity}.

In Algorithm~\ref{alg:Subroutine2}, in order to adapt to the noise parameter $\kappa$, we keep a hierarchical partitioning of the space which becomes more and more refined. This is related to ideas in the continuous bandit literature, in which the goal is to optimize an unknown function over the domain (see \cite{bubeck2011x,munos2011optimistic}). A similar idea was used in~\cite{locatelli2017adaptivity} (for active learning) for the case of smooth regression functions and in~\cite{perchet2013multi} (in the contextual bandit setting) for the case of smooth reward functions, where it is shown that in both these settings, \emph{zooming} strategies lead to natural adaptation to the Tsybakov noise condition.

\section*{Conclusion}
We presented in this work the first adaptive strategy for active learning in the boundary fragment setting, resolving a problem that was open since the formulation of this setting in~\cite{castro2007minimax}, as all known strategies required the knowledge of the characteristic parameters of the problem, which are in general out of reach for practitioners.

\paragraph{Acknowledgement} The work of A. Carpentier and A. Locatelli is supported by the DFG's Emmy Noether grant MuSyAD (CA 1488/1-1).

\bibliographystyle{plain}
\bibliography{library}

\newpage
\begin{appendix}
\section{Proofs}
\subsection{Proof of Theorem~\ref{thm:ls_bound}}
\begin{proof}
Fix $\tilde x \in [0,1]^{d-1}$. In this proof, with a slight abuse of notation, we write $\eta(Z) \doteq \eta((\tilde x, Z))$. Our goal is to find the unique threshold $x_d^* \in [0,1]$ such that we have for any $x_d \geq x^*_d$, ${\eta((\tilde x, x_d)) \geq 1/2}$, and  ${\eta((\tilde x, x_d)) < 1/2}$ for any $x_d < x^*_d$ where $\eta$ is such that Assumption~\ref{asuT} is satisfied for some $\kappa \geq 1$. We will first write the event under which all average estimates used by the algorithm concentrate around their means. For $Z \in [0,1]$ sampled $t \geq 1$ times by the algorithm with $\widehat{\eta}_t(Z) = \sum_{i=1}^t Y_t(Z)$ where $Y_t(Z)$ is the $t$-th observation collected in $(\tilde x, Z)$, consider the event:
\begin{equation*}
|\widehat{\eta}_t(Z) - \eta(Z)| \leq \sqrt{\frac{\log(1/\delta)}{2t}}.
\end{equation*}
By Chernoff-Hoeffding, this event holds with probability at least $1-\delta$. We denote $G_k$ the dyadic grid of $[0,1]$ with step size $2^{-k}$, i.e. $G_k = \{\frac{i}{2^k}, i \in \{1,...,2^k-1\} \}$. Note that there are $2^k-1$ points in $G_k$. Let $K = \ceil{\log_2\left(\frac{1}{2\epsilon}\right)}$, $\delta_k = \frac{\delta}{K 2^{k+1}} $. We define the event $\xi$:
\begin{equation*}
\xi \doteq \Big\{ \forall t,k,i \quad s.t. \quad t \geq 1, k \leq K, Z_{i,k} \in G_k: |\widehat{\eta}_t(Z_{i,k}) - \eta(Z_{i,k})| \leq \sqrt{\frac{\log(\frac{t^2}{\delta_k})}{2t}} \Big\},
\end{equation*}
By a union bound, we have:
\begin{eqnarray*}
\mathbb{P}(\bar{\xi}) & \leq & \sum_{k \leq K} \sum_{Z_{i,k} \in G_k} \sum_{t \geq 1} \frac{\delta_k}{t^2}\\ 
& \leq & \frac{\pi^2}{6} \sum_{k \leq K} \sum_{Z_{i,k} \in G_k} \frac{\delta}{K2^{k+1}}\\
& \leq & \delta,
\end{eqnarray*}
where we use $\sum_{t\geq 1} t^{-2} = \frac{\pi^2}{6} \leq 2$ and the definition of $\delta_k$. This shows that $\mathbb{P}(\xi) \geq 1 - \delta$.\\
Assume that at the beginning of epoch $k$, we have $\Delta_k \doteq R_k - L_k =2^{-k+1}$, and $R_k$ and $L_k$ are such that $x_d^* \in [L_k, R_k]$. As the points $U_k, M_k, V_k$ divide the interval $[L_k, R_k]$ in four subintervals of equal length, and there exists a unique threshold $x_d^* \in [L_k, R_k]$, it implies that there is at most a single point $Z \in \{U_k, M_k, V_k\}$ such that $|Z-x_d^*| < \frac{\Delta_k}{8}$. Consider the case $|U_k - x_d^*| < \frac{\Delta_k}{8}$ - the other cases are handled similarly. We thus have $|M_k - x^*|\geq \frac{\Delta_k}{8}$. This implies by Assumption~\ref{asuT}: 
\begin{equation}\label{eq:bound_eta}
|\eta(M_k) - \frac{1}{2}| \geq c\Big(\frac{\Delta_k}{8}\Big)^{\kappa-1}.
\end{equation}
Without loss of generality, assume that $\widehat{\eta}_{t_k}(M_k) > 1/2$ when the epoch ends for the smallest $t_k$ such that ${|\widehat{\eta}_{t_k}(M_k) - 1/2| \geq 2\sqrt{\frac{\log(t_k/\delta_k)}{2t_k}}}$. On $\xi$, we have:
\begin{equation}\label{eq:bound_eta_hat}
{\eta}(M_k) - \sqrt{\frac{\log(t_k/\delta_k)}{2t_k}} \leq \widehat{\eta}_{t_k}(M_k) \leq {\eta}(M_k)+ \sqrt{\frac{\log(t_k/\delta_k)}{2t_k}}
\end{equation}
Epoch $k$ ends as soon as $\widehat{\eta}_{t_k}(M_k) - 1/2 \geq 2\sqrt{\frac{\log(t_k/\delta_k)}{2t_k}}$. Combining this condition with Equation~\eqref{eq:bound_eta_hat} brings on $\xi$:
\begin{eqnarray*}
2\sqrt{\frac{\log(t_k/\delta_k)}{2t_k}} & \leq & \widehat{\eta}_{t_k}(M_k) - 1/2 \\
& \leq & \eta(M_k) - 1/2 + \sqrt{\frac{\log(t_k/\delta_k)}{2t_k}},
\end{eqnarray*}
which implies that $\eta(M_k) \geq \sqrt{\frac{\log(t_k/\delta_k)}{2t_k}} +1/2 > 1/2$, and we have correctly labeled the point $M_k$ i.e. on $\xi$, $\mathbbm{1}\{\widehat{\eta}_{t_k}(M_k) \geq 1/2 \} = \mathbbm{1}\{\eta(M_k) \geq 1/2\}$. Equations~\eqref{eq:bound_eta_hat} and~\eqref{eq:bound_eta} together yield that the epoch stops if $t_k$ is such that:
\begin{equation}\label{eq:t_k}
3\sqrt{\frac{\log(t_k/\delta_k)}{2t_k}} \leq \eta(M_k) - 1/2,
\end{equation}
implying the following sufficient condition for epoch $k$ to end: $t_k \geq \frac{9\log(t_k/\delta_k)}{2\left(\eta(M_k)-1/2 \right)^2}$. Thus, when the epoch ends we have at most:
$$
t_k \leq \frac{5\log(t_k/\delta_k)}{\left(\eta(M_k)-1/2 \right)^2}.
$$
Denote for now $u=\left(\eta(M_k)-1/2 \right) \leq 1/2$ as $\eta$ is bounded in $[0,1]$ and assume that ${t_k \leq \frac{17\log(1/(u^2\delta_k))}{u^2}}$. Injecting this in Equation~\eqref{eq:t_k} brings that the epoch ends if:
\begin{eqnarray}
t_k & \geq & \frac{5\log(\frac{17\log(1/(u^2\delta_k))}{u^2\delta_k})}{u^2}.
\end{eqnarray}
We now check that $\frac{5\log(17\log(1/(u^2\delta_k))/(u^2\delta_k))}{u^2} \leq \frac{17\log(1/(u^2\delta_k))}{u^2}$. This is true if $5\log(\log(1/(u^2\delta_k)))+5\log(17) \leq 12\log(1/(u^2\delta_k))$. As we have $\delta_k \leq 1$ and $u \leq 1/2$, then $w = 1/(u^2 \delta_k) \geq 4$, and one can easily check that ${5\log(\log(w))+5\log(17) \leq 12\log(w)}$ for any $w \geq 4$.

Using Equation~\eqref{eq:bound_eta}, we thus have the following upper-bound on $t_k$:
$$
t_k \leq \begin{cases}
17c^{-2}\log\left(\frac{1}{c^2\delta_k}\right), \ \text{if} \ \ \kappa = 1,\\

17c^{-2}\log\left(\left(\frac{8}{\Delta_k}\right)^{2(\kappa-1)}\frac{1}{c^2\delta_k}\right)\Big(8 \Delta_k^{-1}\Big)^{2(\kappa-1)}, \ \text{if} \ \ \kappa > 1.
\end{cases}
$$

Similarly, we can show that on $\xi$, we make no mistake in the case $\widehat{\eta}_{t_k}(M_k) < 1/2$ when the epoch stops, and obtain the same bound on $t_k$. Thus on $\xi$ when epoch $k$ ends, we have identified an interval $[L_{k+1}, R_{k+1}]$ of size $\frac{\Delta_k}{2}$ such that $x^* \in [L_{k+1}, R_{k+1}]$. By recurrence, this shows that on $\xi$, we have for any $k \leq K$, $x^* \in [L_k, R_k]$ and $\Delta_k=2^{-k+1}$. We now bound the total budget required for all epochs $k\leq K$ to end on $\xi$. When the algorithm terminates we have requested $N$ labels with the following upper-bound on $N$ for $\kappa > 1$: 
\begin{eqnarray}
N  & = & 3\sum_{k=1}^{K} t_k \nonumber \\
& \leq & 51c^{-2}\sum_{k=1}^{K} \log\left(\left(\frac{8}{\Delta_k}\right)^{2(\kappa-1)}\frac{1}{c^2\delta_k}\right)\Big(8 \Delta_k^{-1}\Big)^{2(\kappa-1)} \nonumber \\
& \leq & 51c^{-2}8^{2(\kappa-1)} \log\left((8 \Delta_K^{-1})^{2(\kappa-1)}\frac{K2^{K+1}}{c^2\delta}\right) \sum_{k=1}^K 2^{2k(\kappa-1)}\label{eq:bound_N} \\
& \leq & 100 c^{-2}8^{2(\kappa-1)}\kappa \log\left(\frac{\log_2(1/\epsilon)}{c^2\delta\epsilon}\right) \sum_{k=1}^K 2^{2k(\kappa-1)} \nonumber \\
& \leq & 100\kappa\log\left(\frac{\log_2(1/\epsilon)}{\epsilon\delta}\right)\log\left(\frac{1}{c}\right)c^{-2}8^{2(\kappa-1)}(\kappa-1)^{-1}(2^{2K(\kappa-1)}-1)\label{eq:bound_N}  \\
& \leq & 200\Big(\log(\frac{1}{\delta})+\log(\frac{1}{\epsilon})\Big) \kappa \log\left(\frac{1}{c}\right) c^{-2}8^{2(\kappa-1)}(\kappa-1)^{-1}\Big(\big(\frac{1}{\epsilon}\big)^{2(\kappa-1)}-1\Big)
\end{eqnarray}
and for $\kappa = 1$:
\begin{eqnarray}
N & \leq & 16\log(1/(c^2\delta_K))c^{-2}K \nonumber \\
& \leq & 64\left(\log\left(\frac{1}{\delta}\right)+\log\left(\frac{1}{\epsilon}\right)\right)\log\left(\frac{1}{c}\right)c^{-2} \log\left(\frac{1}{\epsilon}\right).
\end{eqnarray}
\end{proof}
\subsection{Proof of Theorem~\ref{thm:kappa}}
\begin{proof}
We first define the event $\xi$ on which all the calls to the Subroutine~\ref{alg:Subroutine} are successful. Let $\delta_l = \delta  (\max(1,\floor{\alpha})2^{l(d+1)})^{-1}$. 

\begin{equation*}
\xi \doteq \Big\{ \forall l \geq 1, \forall \tilde a  \in \{0,..., M_{l}\}^{d-1}, |T_{l,\tilde a} - g^*(M_l^{-1} \tilde a)| \leq  \epsilon_l \Big\},
\end{equation*}

At depth $l \geq 1$, we launch $(M_l+1)^{d-1} \leq \max(1,\floor{\alpha})2^{ld}$ line-search instances with confidence parameter $\delta_l$ and precision $\epsilon_l$. Each run, indexed by $\tilde a \in \{0,...,M_l\}^{d-1}$ returns a correct threshold $T_{l,\tilde a}$ along the line segment $\mathcal L_{\tilde a} \doteq \{(M_{l^*}^{-1} \tilde a, x_d), x_d \in [0,1] \}$ such that $|T_{l,\tilde a}-g^*(M_l^{-1} \tilde a)| \leq \epsilon_l$ with probability at least $1-\delta_l$ and using at most $\mathcal O\left(\left(\log(1/\epsilon_l)+\log(1/\delta_l)\right)\epsilon_l^{-2({\kappa-1})}\right)$ samples (see Theorem~\ref{thm:ls_bound}). 

By a union bound, we have ${\mathbb P(\bar{\xi}) \leq \delta \sum_{l\geq 1}2^{-l} \leq 2 \delta}$, which implies that $\mathbb P(\xi) \geq 1 - 2\delta$.\\
At depth $l$, the algorithm performs $(\max(1,\floor{\alpha})2^l +1)^{d-1} \leq \left(2\ceil{\alpha}\right)^{d-1}2^{l(d-1)}$ line-searches. By Equation~\eqref{eq:bound_N} in the proof of Theorem~\ref{thm:ls_bound}, we can upper bound on $\xi$ the total budget that Algorithm~\ref{alg:Subroutine2} uses at depth $l$, with $\epsilon_l = \lambda 2^{-\alpha l} \geq 2^{-\alpha l}$ as $\lambda \geq 1$:
\begin{eqnarray}
N_l & \leq & \left(2\ceil{\alpha}\right)^{d-1}2^{l(d-1)} \log(\frac{2^{l\alpha}}{\delta})200 \log(1/c) c^{-2}(8/\lambda)^{2(\kappa-1)}\frac{\kappa}{\kappa-1} 2^{2l\alpha(\kappa-1)}\\
& \leq & \left(2\ceil{\alpha}\right)^{d-1}200 \log(1/c) c^{-2}(8/\lambda)^{2(\kappa-1)}\frac{\kappa}{\kappa-1} \log(\frac{2^{l\alpha}}{\delta}) 2^{l(2\alpha(\kappa-1)+d-1)}
\end{eqnarray}
We are now ready to bound the minimal depth $l^*$ reached by the algorithm. We also upper-bound naively $l^*$ by $\log_2(n)$, as the budget is insufficient to query all cells once at this depth for $d \geq 2$. We bound the number of samples required to reach depth $l^*$ on $\xi$:
\begin{eqnarray}
\sum_{l=1}^{l^*} N_l & \leq & \sum_{l=1}^{l^*}\left(2\ceil{\alpha}\right)^{d-1}\log(\frac{2^{l\alpha}}{\delta})200 \log(1/c) c^{-2}(8/\lambda)^{2(\kappa-1)}\frac{\kappa}{\kappa-1} 2^{l(2\alpha(\kappa-1)+d-1)} \nonumber \\
& \leq & 200 \left(2\ceil{\alpha}\right)^{d-1}\log(1/c) c^{-2}(8/\lambda)^{2(\kappa-1)}\frac{\kappa}{\kappa-1} \log\left(\frac{2^{l^*\alpha}}{\delta}\right) \sum_{l=1}^{l^*} 2^{l(2\alpha(\kappa-1)+d-1)} \nonumber \\
& \leq & 400 \left(2\ceil{\alpha}\right)^{d-1}\log(1/c) c^{-2}(8/\lambda)^{2(\kappa-1)}\frac{\kappa\alpha}{\kappa-1}\log\left(\frac{n}{\delta}\right) 2^{l^*(2\alpha(\kappa-1)+d-1)}.
\end{eqnarray}
As the algorithm is limited by a maximum budget of $n$ samples, the depth reached on $\xi$ is lower-bounded by the biggest $l^*$ such that:
\begin{equation*}
2^{l^*(2\alpha(\kappa-1)+d-1)} \leq \frac{(\kappa-1)c^2\lambda^{2(\kappa-1)}}{400\left(2\ceil{\alpha}\right)^{d-1} \alpha \log(1/c) \kappa 8^{2(\kappa-1)}} \Big(\frac{n}{\log(n/\delta)}\Big),
\end{equation*}
which implies that a minimum depth:
\begin{equation}\label{eq:L_lb}
l^* \geq \frac{1}{2\alpha(\kappa-1)+d-1}\log_2\left( \frac{(\kappa-1)c^2\lambda^{2(\kappa-1)}}{400\left(2\ceil{\alpha}\right)^{d-1} \alpha \log(1/c) \kappa 8^{2(\kappa-1)}} \Big(\frac{n}{\log(n/\delta)}\Big) \right)-1
\end{equation}
is reached by the algorithm on $\xi$. Let $c_1 = \frac{(\kappa-1)c^2}{400\left(2\ceil{\alpha}\right)^{d-1} \alpha \log(1/c) \kappa 8^{2(\kappa-1)}}$.\\
Let $\tilde a \in \{0, ...,M_{l^*}\}^{d-1}$. On $\xi$, we have:
$$|T_{l^*,\tilde a} - g^*(M_{l^*}^{-1}\tilde a)| \leq \lambda \left(\frac{M_{l^*}}{\max(1, \floor{\alpha})}\right)^{-\alpha}$$
Note that $M_{l^*}$ is a quantity accessible to the algorithm to construct the confidence bands for the estimation of the boundary, as it is simply the step size of the last completed epoch.\\
In what follows, we will consider the threshold estimates $(T_{l^*, \tilde a})_{\tilde a}$ and construct a polynomial approximation of the boundary.

\textbf{Case 1: $\alpha > 1$.}
As in~\cite{castro2007minimax}, we make use of the tensor-product Lagrange polynomials. Let $\tilde q \in \{0, ..., \frac{M_{l^*}}{\floor{\alpha}}-1 \}^{d-1}$ index the cells:
\begin{equation*}
I_{\tilde q} \doteq \Big[\tilde q_1 \floor{\alpha}M_{l^*}^{-1}, (\tilde q_1 +1)\floor{\alpha}M_{l^*}^{-1}\Big] \times ... \times \Big[\tilde q_{d-1} \floor{\alpha}M_{l^*}^{-1}, (\tilde q_{d-1} +1)\floor{\alpha}M_{l^*}^{-1}\Big].
\end{equation*}
These cells partition $[0,1]^{d-1}$ entirely, as we have $M_{l^*}=\floor{\alpha}2^{l^*}$. The tensor-product Lagrange polynomials are defined as follows:
\begin{equation*}
Q_{\tilde q, \tilde a}(\tilde x) \doteq \prod_{i=1}^{d-1} \ \prod_{\substack{ 0 \leq j \leq \floor{\alpha} \\ j \neq \tilde a_i - \floor{\alpha} \tilde q_i}} \frac{\tilde x_i-M_{l^*}^{-1}(\floor{\alpha}\tilde q_i+j)}{M_{l^*}^{-1}\tilde a_i - M_{l^*}^{-1}(\floor{\alpha}\tilde q_i+j)}.
\end{equation*}
It is easily shown that (\cite{castro2007minimax,castro2007active}):
\begin{equation}\label{eq:bound_Q}
\max_{\tilde x \in I_{\tilde q}} |Q_{\tilde a, \tilde q}(\tilde x)| \leq \floor{\alpha}^{(d-1)\floor{\alpha}}.
\end{equation}
The tensor-product Lagrange polynomial interpolation of $g^*$ for $\tilde x \in I_{\tilde q}$ is:
\begin{equation}
P_{\tilde q}(\tilde x) = \sum_{\tilde a: M_{l^*}^{-1}\tilde a \in I_{\tilde q}} g^*(M_{l^*}^{-1} \tilde a)Q_{\tilde q, \tilde a}(\tilde x)
\end{equation}
and we define the polynomial interpolation of $g^*$ for $\tilde x \in I_{\tilde q}$:
\begin{equation}
\widehat{P}_{\tilde q}(\tilde x) = \sum_{\tilde a: M_{l^*}^{-1}\tilde a \in I_{\tilde q}} T_{l^*, \tilde a}Q_{\tilde q, \tilde a}(\tilde x).
\end{equation}
On $\xi$, since $\epsilon_{l^*} = \Big(\frac{M_{l^*}}{\floor{\alpha}}\Big)^{-\alpha}$:
\begin{equation}\label{bound:R}
|T_{l^*, \tilde a} - g^*(M_{l^*}^{-1} \tilde a)| \leq \lambda \Big(\frac{M_{l^*}}{\floor{\alpha}}\Big)^{-\alpha}.
\end{equation}
For any $\tilde x \in I_{\tilde q}$, the previous equation brings on $\xi$:
\begin{eqnarray}\label{eq:var}
|\widehat{P}_{\tilde q}(\tilde x) - P_{\tilde q}(\tilde x)| & =  &\big\lvert \sum_{\tilde a: M_{l^*}^{-1}\tilde a \in I_{\tilde q}} \Big(T_{l^*, \tilde a} - g^*(M_{l^*}^{-1} \tilde a)\Big)Q_{\tilde q, \tilde a}(\tilde x) \big\rvert \nonumber \\
& \leq &  \sum_{\tilde a: M_{l^*}^{-1}\tilde a \in I_{\tilde q}} \lambda \Big(\frac{M_{l^*}}{\floor{\alpha}}\Big)^{-\alpha}  \big\lvert Q_{\tilde q, \tilde a}(\tilde x) \big\rvert \nonumber \\
& \leq &  \sum_{\tilde a: M_{l^*}^{-1}\tilde a \in I_{\tilde q}} \lambda \Big(\frac{M_{l^*}}{\floor{\alpha}}\Big)^{-\alpha}  \floor{\alpha}^{(d-1)\floor{\alpha}} \nonumber \\
& \leq &  \ceil{\alpha}^{d-1}\floor{\alpha}^{(d-1)\floor{\alpha}} \floor{\alpha}^{\alpha}\lambda M_{l^*}^{-\alpha} \nonumber \\
& \leq & \ceil{\alpha}^{d\ceil{\alpha}} \lambda M_{l^*}^{-\alpha},
\end{eqnarray}
where we use Equation~\eqref{eq:bound_Q} in line 4, and upper-bound the number of terms in the sum by $\ceil{\alpha}^{d-1}$.\\
We now turn our attention to the approximation properties of $P_{\tilde q}$ with respect to $g^*$, which do not depend on $\xi$. For any $\tilde x \in I_{\tilde q}$ and $g^* \in \Sigma(\lambda, \alpha)$, we have:
\begin{eqnarray}
|P_{\tilde q}(\tilde x) - g^*(\tilde x)| & = & |P_{\tilde q}(\tilde x) - \text{TP}_{\tilde q\floor{\alpha}M_{l^*}^{-1}}(\tilde x) + \text{TP}_{\tilde q\floor{\alpha}M_{l^*}^{-1}}(\tilde x) - g^*(\tilde x)|\nonumber \\
& \leq & |P_{\tilde q}(\tilde x) - \text{TP}_{\tilde q\floor{\alpha}M_{l^*}^{-1}}(\tilde x)| + |\text{TP}_{\tilde q\floor{\alpha}M_{l^*}^{-1}}(\tilde x) - g^*(\tilde x)|\nonumber \\
& \leq & |P_{\tilde q}(\tilde x) - \text{TP}_{\tilde q\floor{\alpha}M_{l^*}^{-1}}(\tilde x)| + \lambda \Big(\frac{M_{l^*}}{\floor{\alpha}}\Big)^{-\alpha}\label{eq:approx},
\end{eqnarray}
where $\text{TP}_x$ is the Taylor polynomial expansion of $g$ in $x$ of degree $\floor{\alpha}$. As the Taylor polynomial expansion is of degree $\floor{\alpha}$, it is also possible to write $\text{TP}_{\tilde q\floor{\alpha}M_{l^*}^{-1}}$ in the tensor-product Lagrange polynomials basis, bringing: 
\begin{eqnarray*}
|P_{\tilde q}(\tilde x) - \text{TP}_{\tilde q\floor{\alpha}M_{l^*}^{-1}}(\tilde x)| & = & \big\lvert \sum_{\tilde a: M_{l^*}^{-1}\tilde a \in I_{\tilde q}} \Big( g^*(M_{l^*}^{-1} \tilde a)-\text{TP}_{\tilde q\floor{\alpha}M_{l^*}^{-1}}(M_{l^*}^{-1}\tilde a)\Big)Q_{\tilde q, \tilde a}(\tilde x) \big\rvert \\
& \leq &  \sum_{\tilde a: M_{l^*}^{-1}\tilde a \in I_{\tilde q}} |g^*(M_{l^*}^{-1} \tilde a)-\text{TP}_{\tilde q\floor{\alpha}M_{l^*}^{-1}}(M_{l^*}^{-1}\tilde a)| |Q_{\tilde q, \tilde a}(\tilde x)| \big\rvert \\
& \leq &  \sum_{\tilde a: M_{l^*}^{-1}\tilde a \in I_{\tilde q}} \lambda \Big(\frac{M_{l^*}}{\floor{\alpha}}\Big)^{-\alpha}  \big\lvert Q_{\tilde q, \tilde a}(\tilde x) \big\rvert \\
& \leq &  \sum_{\tilde a: M_{l^*}^{-1}\tilde a \in I_{\tilde q}} \lambda \Big(\frac{M_{l^*}}{\floor{\alpha}}\Big)^{-\alpha}  \floor{\alpha}^{(d-1)\floor{\alpha}}\\
& \leq &  \ceil{\alpha}^{d-1}\floor{\alpha}^{(d-1)\floor{\alpha}} \floor{\alpha}^{\alpha}\lambda M_{l^*}^{-\alpha}\\
& \leq & \ceil{\alpha}^{d\ceil{\alpha}} \lambda M_{l^*}^{-\alpha},
\end{eqnarray*}
where the third line is obtained by using Assumption~\ref{asuH} as $g^*$ is $\alpha$-smooth. Combining this with Equation~\eqref{eq:approx} yields the following inequality:
\begin{eqnarray}\label{eq:bias}
|P_{\tilde q}(\tilde x) - g^*(\tilde x)| & \leq & 2\ceil{\alpha}^{d\ceil{\alpha}} \lambda M_{l^*}^{-\alpha}.
\end{eqnarray}
We are now ready to conclude the proof. Combining Equations~\eqref{eq:var} and~\eqref{eq:bias} allows us to write:
\begin{eqnarray*}
|\widehat{P}_{\tilde q}(\tilde x) - g^*(\tilde x)| & \leq & |\widehat{P}_{\tilde q}(\tilde x) - P_{\tilde q}(\tilde x)| + |P_{\tilde q}(\tilde x) - g^*(\tilde x)|\\
& \leq & 3\ceil{\alpha}^{d\ceil{\alpha}} \lambda M_{l^*}^{-\alpha},
\end{eqnarray*}
which brings immediately with $b_{l^*} = \ceil{\alpha}^{d\ceil{\alpha}} \lambda M_{l^*}^{-\alpha}$ as defined in the algorithm:
\begin{equation*}
0 < b_{l^*} \leq (\widehat{P}_{\tilde q}(\tilde x)+ 4b_{l^*}) - g^*(\tilde x) \leq 7b_{l^*}.
\end{equation*}
This implies directly the following inclusions on $\xi$:
\begin{equation*}
\{x: x_d \geq g^*(\tilde x) + 7 b_{l^*} \} \subset S^1 \subset \{x: x_d > g^*(\tilde x)\}
\end{equation*}
Through similar considerations, it is easily shown that on $\xi$, we also have:
\begin{equation*}
\{x: x_d \leq g^*(\tilde x) - 7 b_{l^*} \} \subset S^0 \subset \{x: x_d < g^*(\tilde x)\}.
\end{equation*}

This shows that the procedure is $(n, \delta, \Delta_{l^*})$-correct with:
\begin{equation*}
\Delta_{l^*} \leq 7\ceil{\alpha}^{d\ceil{\alpha}} 2^\alpha \lambda^{\frac{d-1}{2\alpha(\kappa-1)+d-1}} \Big(\frac{\log(n/\delta)}{c_1n} \Big)^{\frac{\alpha}{2\alpha(\kappa-1)+d-1}}.
\end{equation*}

\textbf{Case 2: $\alpha \leq 1$.} We simply use a constant approximation directly on the cells:
\begin{equation*}
C_{\tilde h} \doteq \Big[\tilde h_1 M_{l^*}^{-1}, (\tilde h_1 +1)M_{l^*}^{-1}\Big] \times ... \times \Big[\tilde h_{d-1} M_{l^*}^{-1}, (\tilde h_{d-1} +1)M_{l^*}^{-1}\Big],
\end{equation*}
indexed by $\tilde h \in \{0, ..., M_{l^*}-1 \}$. For $\alpha \leq 1$, the assumption on the smoothness of the boundary simply yields for any $\tilde h \in \{0, ..., M_{l^*}-1 \}$ and any $\tilde x, \tilde y \in C_{\tilde h}$:
\begin{equation}\label{eq:alpha_one}
|g^*(\tilde x) - g^*(\tilde y)| \leq \lambda ||\tilde x - \tilde y||_{\infty}^\alpha \leq \lambda M_{l^*}^{-\alpha}.
\end{equation}
Note that for $\alpha \leq 1$, we have $b_{l^*} = \lambda M_{l^*}^{-\alpha}$, as we have $\ceil{\alpha} = 1$. Equation~\eqref{bound:R} and Equation~\eqref{eq:alpha_one} yield for any $\tilde x \in C_{\tilde h}$:
\begin{equation*}
0 < b_{l^*} \leq T_{l^*, \tilde h} +2 b_{l^*} - g^*(\tilde x) \leq 4 b_{l^*},
\end{equation*}
which shows the $(n, \delta, \Delta_{l^*})$ correctness of the procedure with:
\begin{equation*}
\Delta_{l^*} \leq 2^{\alpha} 5\lambda^{\frac{d-1}{2\alpha(\kappa-1)+d-1}} \Big(\frac{\log(n/\delta)}{c_1n} \Big)^{\frac{\alpha}{2\alpha(\kappa-1)+d-1}}
\end{equation*}
\end{proof}
\subsection{Proof of Proposition~\ref{thm_adaptive} and Theorem~\ref{thm:adap_rate}}\label{proof:adap_rate}
\begin{proof}
The proof follows from arguments in~\cite{locatelli2017adaptivity}, adapted to this different notion of correctness.

Set as in Algorithm~\ref{alg:sa}:
$$n_0 = \frac{n}{\floor{\log(n)}^2},~~~\delta_0 = \frac{\delta}{\floor{\log(n)}^2},~~~\text{and}~~\alpha_i = \frac{i}{\floor{\log(n)}}.$$

In Algorithm~\ref{alg:sa}, the Subroutine is launched $\floor{\log(n)}^2$ times on $\floor{\log(n)}^2$ independent subsamples of size $n_0$. We index each launch by $i$, which corresponds to the launch with smoothness parameter $\alpha_i$. Let $i^*$ be the largest integer $1\leq i\leq \floor{\log(n)}^2$ such that $\alpha_i \leq \alpha$.

Since the Subroutine is strongly $(\delta_0,\Delta_{\alpha}, n_0)$-correct for any $\alpha \in [\floor{\log(n)}^{-1}, \floor{\log(n)}]$, it holds by Definition~\ref{def:correct} that for any $i \leq i^*$, with probability larger than $1-\delta_0$
$$\Big\{x\in [0,1]^d : x_d - g^*(\tilde x) > \Delta_{\alpha_i}\Big\} \subset S^1_i \subset \Big\{x\in [0,1]^d : x_d - g^*(\tilde x) > 0 \Big\}
$$
and
$$
\Big\{x\in [0,1]^d :g^*(\tilde x) - x_d > \Delta_{\alpha_i}\Big\} \subset S^0_i \subset \Big\{x\in [0,1]^d : g^*(\tilde x) - x_d > 0 \Big\}.$$
So by an union bound we know that with probability larger than $1-\floor{\log(n)}^2\delta_0 = 1-\delta$, the above equations hold jointly for any $i \leq i^*$.

This implies that with probability larger than $1-\delta$, we have for any $i' \leq i \leq i^*$, and for any $y\in \{0,1\}$, that
$$S^{y}_i \cap s^{1-y}_{i'} = \emptyset,$$
i.e.~the labeled regions of $[0,1]^d$ are not in disagreement for any two runs of the algorithm that are indexed with parameters smaller than $i^*$. So we know that just after iteration $i^*$ of Algorithm~\ref{alg:sa}, we have with probability larger than $1-\delta$, that for any $y\in \{0,1\}$
$$\bigcup_{i \leq i^*} S^{y}_i \subset s^{y}_{i^*}.$$
Since the sets $s^{y}_i$ are strictly growing but disjoint with the iterations $i$ by definition of Algorithm~\ref{alg:sa} (i.e.~$s_i^{k} \subset s_{i+1}^{k}$ and $s_i^{k} \cap s_{i}^{1-k} = \emptyset$), it holds in particular that with probability larger than $1-\delta$ and for any $y\in \{0,1\}$
$$\bigcup_{i \leq i^*} S^{y}_i \subset s^{y}_{\floor{\log(n)}^2}~~~\text{and}~~~s_{\floor{\log(n)}^2}^{y} \cap s_{\floor{\log(n)}^2}^{1-y} = \emptyset.$$
This finishes the proof of Proposition~\ref{thm_adaptive}.\\

By Proposition~\ref{thm_adaptive}, Algorithm~\ref{alg:sa} is weakly-$(\delta_0, \Delta_{\alpha_i}, n_0)$ correct for the largest $i$ such that $\alpha_i \leq \alpha$, with $\Delta_{\alpha_i}$ bounded as:
\begin{equation*}
\Delta_{\alpha_i} \leq 7\ceil{\alpha_i}^{d\ceil{\alpha_i}} 2^{\alpha_i} \lambda \Big(\frac{\log^3(n/\delta)}{c_1n} \Big)^{\frac{\alpha_i}{2\alpha_i(\kappa-1)+d-1}},
\end{equation*}
with $c_1 = \frac{(\kappa-1)c^2}{400\left(2\ceil{\alpha}\right)^{d-1} \alpha \log(1/c) \kappa 8^{2(\kappa-1)}}$.\\
By definition of $\alpha_i$, which is on a grid of step $\floor{\log(n)}^{-1}$, we have:
\begin{equation*}
\alpha - \frac{1}{\floor{\log(n)}} \leq \alpha_i \leq \alpha.
\end{equation*}
This yields for the exponent in the rate:
\begin{equation*}
-\frac{\alpha_i}{2\alpha_i(\kappa-1)+d-1} \leq -\frac{\alpha}{2\alpha(\kappa-1)+d-1}+ \frac{\floor{\log(n)}^{-1}}{2\alpha(\kappa-1)+d-1}.
\end{equation*}
The result follows by noticing that:
\begin{eqnarray*}
n^{\frac{1}{\floor{\log(n)}(2\alpha(\kappa-1)+d-1)}} \leq \exp\left(\frac{\log(n)}{\floor{\log(n)}(d-1)}\right)
\end{eqnarray*}
and thus this term only affects the rate as a multiplicative constant that does not depend on $n, \delta$ and $\lambda$.
\end{proof}

\subsection{Proof of Theorem~\ref{thm:lb}}\label{proof:lb}
\begin{proof}
The basic argument is based on standard applications of Fano's inequality, in particular on a useful form given in Theorem 2.5 in~\cite{tsybakovintroduction} (which we recall hereunder). The main work is in constructing a suitable family of problems satisfying the conditions of Theorem~\ref{thm:tsy} and matching our distributional requirements.

\begin{thm}[Tsybakov]\label{thm:tsy} Let $\mathcal{H}$ be a class of models, $d: \mathcal{H} \times \mathcal{H} \rightarrow \mathbb R^+$ a pseudo-metric, and $\{P_{\eta}, \eta \in \mathcal{H}\}$ a collection of probability measures associated with $\mathcal{H}$. Assume there exists a subset $\{\eta_0, ..., \eta_M \}$ of $\mathcal{H}$ such that:
\begin{enumerate}
\item $d(\eta_i, \eta_j) \geq 2s > 0$ for all $0 \leq i < j \leq M$
\item $P_{\eta_i}$ is absolutely continuous with respect to $P_{\eta_0}$ for every $0 < i \leq M$
\item $\frac{1}{M} \sum_{i=1}^M \textsc{KL}(P_{\eta_i}, P_{\eta_0}) \leq \alpha \log(M)$, for $0 < \alpha < \frac{1}{8}$
\end{enumerate}
then
$$
\inf_{\hat \eta} \sup_{\eta \in \mathcal{H}} P_\eta \big(d(\hat \eta, \eta) \geq s \big) \geq \frac{\sqrt{M}}{1+\sqrt{M}}\Big(1-2\alpha-\sqrt{\frac{2\alpha}{\log(M)}} \Big),
$$
where the infimum is taken over all possible estimators of $\eta$ based on a sample from $P_\eta$.
\end{thm}

Let $\alpha > 0$ and $d \in \mathbb{N}$, $d > 1$. For $x\in \mathbb R^d$, we write $x = (x^{(1)}, \cdots, x^{(d)})$ and $x^{(i)}$ denotes the value of the $i$-th coordinate of $x$. As previously, for $x \in [0,1]^d$, we use the notation $\tilde x = (x^{(1)}, \ldots, x^{(d-1)})$.

Consider the grid of $[0,1]^{d-1}$ of step size $2 \Delta^{1/\alpha}$, $\Delta > 0$. There are 
$$K = 2^{1-d}\Delta^{(1-d)/\alpha},$$
disjoint hypercubes in this grid, and we write them $(H'_k)_{k \leq K}$. For $k \leq K$, let $\tilde x_k$ be the barycenter of $H_k'$.

We now define the partition of $[0,1]^d$ :
$$[0,1]^d = \bigcup_{k=1}^K H_k = \bigcup_{k=1}^K (H_k'\times [0,1]),$$
where $H_k = (H_k'\times [0,1])$ is an hyper-rectangle corresponding to $H_k'$ - these are hyper-rectangles of side $2 \Delta^{1/\alpha}$ along the first $(d-1)$ dimensions, and side $1$ along the last dimension.\\


We define $f$ for any $z\in[\frac{1}{2}\Delta^{1/\alpha},\Delta^{1/\alpha}]$ as
\begin{equation*}
    f(z)=
    \begin{cases}
             C_{\lambda, \alpha}4^{\alpha - 1}\Big(\Delta^{1/\alpha}-z\Big)^\alpha, & \text{if}\ \frac{3}{4}\Delta^{1/\alpha} <z \leq \Delta^{1/\alpha}\\
       C_{\lambda, \alpha}\Big(\frac{\Delta}{2} - 4^{\alpha - 1} \big(z - \frac{1}{2}\Delta^{1/\alpha} \big)^\alpha\Big), & \text{if}\  \frac{1}{2}\Delta^{1/\alpha} \leq z \leq \frac{3}{4}\Delta^{1/\alpha},
    \end{cases}
\end{equation*}
where $C_{\lambda, \alpha}>0$ is a small constant that depends only on $\alpha, \lambda$.

For $k \leq K$, and for any $\tilde x \in H'_k$, we write
\begin{equation*}
    \Psi_{k}(\tilde x)
    \begin{cases}
       \frac{C_{\lambda, \alpha}\Delta}{2}, & \quad \text{if}\ \quad |\tilde x - \tilde x_k|_2 \leq \frac{\Delta^{1/\alpha}}{2} \\
              0, & \text{if}\quad |\tilde x - \tilde x_k|_2 \geq \Delta^{1/\alpha}\\
       \ f(|\tilde x - \tilde x_k|), & \text{otherwise},
    \end{cases}
\end{equation*}
which we use to define $g_{k_s}$ over the same domain, for $s \in \{-1,1\}$:
\begin{equation*}
    g_{k,s}(\tilde x)= \frac{1}{2} + s \Psi_{k}(\tilde x)
\end{equation*}

$f$ is such that $f(\frac{1}{2}\Delta^{1/\alpha})) = \frac{C_{\lambda, \alpha}\Delta}{2}$, and $f(\Delta^{1/\alpha}) = 0$. Moreover, it is $(\lambda,\alpha)$-H\"older on $[\frac{1}{2}\Delta^{1/\alpha},\Delta^{1/\alpha}]$ for $C_{\lambda, \alpha}$ small enough (depending only on $\alpha, \lambda$), and such that all its derivatives are $0$ in $\frac{1}{2}\Delta^{1/\alpha}$, $\Delta^{1/\alpha}$. By definition of $\Psi_{k,s}$, it holds that $g_{k,s}$ is in $\Sigma(\lambda, \alpha)$ restricted to $H'_k$.

We now define $\eta_{k,s}$ for $x \in H_k$:
\begin{equation*}
    \eta_{k,s}(x)=
    \begin{cases}
       c |x_d - g_{k,s}(\tilde x)+ 2\Psi_{k}(\tilde x)|^{\kappa-1} & \text{if}\quad s(x_d - g_{k,s}(\tilde x)) > 2 \Psi_k(\tilde x) \\
       c |x_d - g_{k,s}(\tilde x)|^{\kappa-1} & \text{otherwise.}
    \end{cases}
\end{equation*}

We see immediately by definition of $\eta_{k,s}$ that it satisfies Assumption~\ref{asuT}, and that $\eta_{k,-1}(x) = \eta_{k, 1}(x)$ for $\{x: |x_d - 1/2| \geq \Psi_k(\tilde x) \}$ (i.e. $\eta_{k,s}$ only depends on $s$ in a small band around the decision boundary).

For $\sigma \in \{-1,1\}^K$, we define for any $\tilde x \in [0,1]^{d-1}$ the function
$$g^*_\sigma(\tilde x) = \sum_{k \leq K} g_{k, \sigma_k} (\tilde x) \mathbf 1\{\tilde x \in H'_k\}.$$
Note that since each $g_{k,s}$ is in $\Sigma(\lambda, \alpha)$ restricted to $H'_k$, and by definition of $g_{k,s}$ at the borders of each $H'_k$, it holds that $g^*_\sigma$ is in $\Sigma(\lambda, \alpha)$ on $[0,1]^{d-1}$.\\


We now define the marginal distribution $\mathbb P_X$ of $X$. To simplify notations, we first define for any $x \in H_k$: $D_k(x) = \min(|x_d - g_{k,1}(\tilde x)|, |x_d - g_{k,-1}(\tilde x)|)$ and $D(x) = \sum_{k=1}^K D_k(x) \mathbf 1\{x \in H_k\}$. This is simply the distance from $x$ to the closest possible location of the boundary, and it does not depend on $s \in \{-1, 1\}$. We define $p_k$ for $x \in H_k$ for $\kappa' > \kappa - 1$:
$$
p_k(x)=\begin{cases}
	C_1 D_k(x)^{\kappa'-\kappa} &~~\text{if}~~D_k(x) \leq \Delta_0\\
    C_2&~~\text{otherwise}.
\end{cases}
$$
This allows us to define the density:
$$
p(x) = \sum_{k=1}^K p_k(x) \mathbf 1\{x \in H_k\},
$$
where the constants $C_1$ and $C_2$ are chosen such that Assumption~\ref{asuM} is satisfied and $p$ integrates to $1$ over $[0,1]^d$.\\

Finally, for any $\sigma \in  \{-1, +1\}^K$, we define $P_{\eta_\sigma}$ as the measure of the data in our setting when the density of $\mathbb P_X$ is $p$, and where the regression function $\mathbb P_{Y|X}$ providing the distribution of the labels is $\eta_\sigma$. By a slight abuse of notation, we write $P_\sigma = P_{\eta_\sigma}$. We write 
$$\mathcal{H} = \{\eta_{\sigma}: \sigma \in \{-1, +1\}^K \}.$$
For any element $\eta_\sigma$ of $\mathcal H$, $P_\sigma$ satisfies Assumptions, ~\ref{asuH},~\ref{asuT} and ~\ref{asuM} by construction.\\

We define $P_{\sigma,n}$ the joint distribution $(X_i,Y_i)_{i=1}^n$ of samples collected by any (possibly active) fixed sampling strategy $\Pi_n$ under $P_\sigma$, where $\Pi_n = \{\pi_i\}_{i \leq n}$, and $\pi_t(x,\{(X_i, Y_i)\}_{i<t})$ is the sampling strategy at time $t$ that depends on the samples collected up to time $t$. $\pi_t$ defines the sampling rule $\pi_t (x, \{(X_i, Y_i)\}_{i < t}) = P_{\pi, \sigma}(X_t = x| (X_1, Y_1), \ldots , (X_{t-1}, Y_{t-1}))$, for any $x \in [0,1]^d$. We remark here that this sampling mechanism may depend on $\mathbb P_X$, which is why we have constructed $\mathbb P_X$ such that it does not depend on $\sigma$. This is crucial for Proposition~\ref{prop:castro} (from \cite{castro2007minimax}) to hold. As $\mathbb P_X$ does not depend on $\sigma$, we have immediately that $\forall i \leq M$, $P_{\sigma_i,n}$ is absolutely continuous with respect to $P_{\sigma_0,n}$.

\begin{prop}[Gilbert-Varshamov]\label{prop:gilb} For $K \geq 8$ there exists a subset $\{\sigma_0, ..., \sigma_M \} \subset \{-1, 1\}^K$ such that $\sigma_0 = \{1, ..., 1\}$, $\rho(\sigma_i, \sigma_j) \geq \frac{K}{8}$ for any $0 \leq i  < j \leq M$ and $M \geq 2^{K/8}$, where $\rho$ stands for the Hamming distance between two sets of length $K$.
\end{prop}

We denote $\mathcal{H'} \doteq \{\eta_{\sigma_0}, \cdots ,\eta_{\sigma_M}\}$ a subset of $\mathcal{H}$ of cardinality $M \geq 2^{K/8}$ with $K \geq 8$ such that for any $1 \leq k < j \leq M$, we have $\rho(\sigma_k, \sigma_j) \geq K/8$. We know such a subset exists by Proposition~\ref{prop:gilb}.


\begin{prop}[Castro and Nowak]\label{prop:castro} For any $\sigma \in \mathcal{H}$ such that $\sigma \neq \sigma_0$ and  $\Delta$ small enough such that $\eta_{\sigma}, \eta_{\sigma_0}$ take values only in $[1/5, 4/5]$ and $\mathbb P_X$ does not depend on $\sigma$, we have:
\begin{eqnarray*}
\textsc{KL}( P_{\sigma, n} ||  P_{\sigma_0, n}) & \leq & 7n \max_{x\in [0,1]^d}(\eta_\sigma(x) - \eta_{\sigma_0}(x))^2.
\end{eqnarray*}
where $\textsc{KL}(. || . )$ is the Kullback-Leibler divergence between two-distributions, and $P_{\sigma,n}$ stands for the joint distribution $(X_i,Y_i)_{i=1}^n$ of samples collected by any (possibly active) fixed sampling strategy under $P_\sigma$. 
\end{prop}
This proposition is a consequence of the analysis in~\cite{castro2008minimax} (Theorem 1 and 3, and Lemma 1). A proof can be found in~\cite{minsker2012a}.\\

By Definition of the $\eta_\sigma$, we know that $\max_{x\in [0,1]^d}|\eta_\sigma(x) - \eta_{\sigma_0}(x)| \leq c (2C_{\lambda, \alpha} \Delta)^{\kappa-1}$, and so Proposition~\ref{prop:castro} implies that for any $\sigma \in \mathcal{H}'$:
\begin{eqnarray*}\label{kl_lb}
\textsc{KL}( P_{\sigma, n} ||  P_{\sigma_0, n}) & \leq & 7n \max_{x\in [0,1]^d}(\eta_\sigma(x) - \eta_{\sigma_0}(x))^2\\
& \leq & 7n c^2 (2C_{\lambda, \alpha})^{2(\kappa-1)} \Delta^{2(\kappa-1)}.
\end{eqnarray*}
  So we have :
\begin{equation*}
\frac{1}{M}\sum_{\sigma \in \mathcal{H}'} \textsc{KL}( P_{\sigma, n} ||  P_{\sigma_0, n}) \leq 7n c^2 (2C_{\lambda, \alpha})^{2(\kappa-1)} \Delta^{2(\kappa-1)} < \frac{K}{8^2} \leq \frac{\log(|\mathcal{H}'|)}{8},
\end{equation*}
for $n$ larger than a constant that depends only on $\alpha, \lambda$, and setting
\begin{equation*}
\Delta = C_3 n^{-\alpha/(2(\kappa-1)\alpha + d - 1)},
\end{equation*}
as $K = C_4 \Delta^{(1-d)/\alpha}$. This implies that for this choice of $\Delta$, the third condition in Theorem~\ref{thm:tsy} is satisfied.

Finally, we define the pseudo-metric as follows:
$$
d(\eta,\eta') = \int_{x \in [0,1]^d} \mathbf 1\{\mathrm{sign}(\eta(x)-1/2) \neq \mathrm{sign}(\eta'(x)-1/2)\}D(x)^{\kappa-1}p(x)\text{d}x.
$$
For $\sigma, \sigma' \in \mathcal H'$, we have:
\begin{eqnarray*}
d(\eta_\sigma, \eta_{\sigma'}) & = & \int_{x \in [0,1]^d} \mathbf 1\{\mathrm{sign}(\eta_{\sigma}(x)-1/2) \neq \mathrm{sign}(\eta_{\sigma'}(x)-1/2)\}D(x)^{\kappa'-1}\text{d}x.\\
& = & C_5 \rho(\sigma, \sigma') \int_{\tilde x \in H'_1} \left(\int_{|x_d-1/2| \leq \Psi_1(\tilde x)} \min(|x_d - g_{1,1}(\tilde x)|,|x_d - g_{1,-1}(\tilde x)|)^{\kappa'-1}\text{d}x_d \right) \text{d}\tilde x\\
& = & 2C_5 \rho(\sigma, \sigma') \int_{\tilde x \in H'_1} \left( \int_{1/2 }^{1/2+\Psi_k(\tilde x)} \left|x_d - g_{1,1}(\tilde x)\right|^{\kappa'-1}\text{d}x_d \right) \text{d} \tilde x\\
& \geq & 2C_5 \rho(\sigma, \sigma') \int_{|\tilde x-x_k|_2 \leq \frac{\Delta^{1/\alpha}}{2}} \left( \int_{1/2 }^{1/2+\frac{C_{\lambda,\alpha}\Delta}{2}} \left|x_d - \frac{C_{\lambda,\alpha}\Delta}{2} + \frac{1}{2}\right|^{\kappa'-1}\text{d}x_d \right) \text{d} \tilde x\\
& \geq & C_6 \rho(\sigma, \sigma')\Delta^{(d-1)/\alpha}\Delta^{\kappa'}\\
& \geq & C_7 \Delta^{\kappa'},
\end{eqnarray*}
where we use the definition of $p$ in the first line, the definition of $\eta_\sigma$ and $\rho(\sigma, \sigma')$ and Fubini's theorem in the second line, and the lower bound on $\rho(\sigma, \sigma')$ by definition of $\mathcal H'$ in the last line.\\

All assumptions in Theorem~\ref{thm:tsy} are thus satisfied with $s = C_7 \Delta^{\kappa'}$ and $\Delta = C_3n^{-\alpha/(2(\kappa-1)\alpha+d-1)}$. For any $\eta_\sigma \in \mathcal H'$, and any $\hat \eta: [0,1]^d \rightarrow [0,1]$:
\begin{eqnarray*}
d(\hat \eta_n, \eta_\sigma) & = & \int_{x \in [0,1]^d} \mathbf 1\{\mathrm{sign}(\hat \eta_n(x)-1/2) \neq \mathrm{sign}(\eta_\sigma(x)-1/2)\}D(x)^{\kappa-1}p(x)dx\\
& \leq & c^{-1} \int_{x \in [0,1]^d} \mathbf 1\{\mathrm{sign}(\hat \eta_n(x)-1/2) \neq \mathrm{sign}(\eta_\sigma(x)-1/2)\}|1-2\eta_\sigma(x)|p(x)dx\\
& \leq & \frac{R_{P_\sigma}(\hat \eta_n) - R_{P_\sigma}(\eta_\sigma)}{c}
\end{eqnarray*}
where we use in the second line the fact that $\eta_\sigma$ satisfies Assumption~\ref{asuT} with constant $c$, and thus under $P_\sigma$, we have $d(\hat \eta, \eta_\sigma) \leq  c^{-1} \mathcal E_{P_\sigma}(\hat \eta)$. We can now apply Theorem~\ref{thm:tsy}, which yields for any fixed sampling strategy $\pi_n$ as defined previously:
$$
\inf_{\hat \eta_n} \sup_{\eta_\sigma \in \mathcal{H}} P_{\sigma,n} \left( \mathcal{E}(\hat \eta_n)\geq C_8 n^{-\kappa'\alpha/(2\alpha(\kappa-1)+d-1)} \right) \geq C_9,
$$
where $C_9$ is a small universal constant. We conclude by applying Markov's inequality, and taking the infimum over (possibly active) sampling strategies $\Pi_n$ (as this holds for any strategy $\Pi_n$).
\end{proof}

\subsection{Proof of Theorem~\ref{thm:lb_passive} (passive lower bound)}\label{proof:lb_passive}

\begin{proof}
In the passive setting, the proof is the same but we need a different bound on the quantity:
$$\text{KL}(P_{\sigma,n} || P_{\sigma_0,n}) = n \int_{x: \eta_\sigma(x) \neq \eta_{\sigma_0}(x)} d_\text{KL}(\eta_{\sigma}(x),\eta_{\sigma_0}(x))p(x)\text{d}x,$$
where $d_\text{KL}(p,q)$ stands for the Kullback-Leibler divergence between two Bernoulli distributions of parameters $p,q$. Instead, we bound it as:
$$
\text{KL}(P_{\sigma,n} || P_{\sigma_0,n}) \leq n C_{10} \Delta^{\kappa'+\kappa-1},
$$
using $d_\text{KL}(\eta_{\sigma}(x),\eta_{\sigma_0}(x)) \leq C_{11} \Delta^{2(\kappa-1)}$ by Pinsker's inequality for $\eta(x) \in [1/5,4/5]$, and the definition of $p(x)$. We conclude by setting $\Delta = C_{12}n^{-\alpha/(\alpha(\kappa+\kappa'-1)+d-1)}$ to satisfy the assumptions of Theorem~\ref{thm:tsy}.
\end{proof}
\end{appendix}
\end{document}